%% file: main.tex
\documentclass{CUP-JNL-EDS}%%option ruler --> print line number in the margin

\usepackage{latexsym}
\usepackage{graphicx}
\usepackage{float}
\usepackage{multicol,multirow}
\usepackage{amsmath,amssymb,amsfonts}
\usepackage{mathrsfs}
\usepackage{amsthm}
\usepackage{rotating}
\usepackage{appendix}
\usepackage[natbib,
            style=apa,
            uniquename=false,
            giveninits=false,
            uniquelist=false
            ]{biblatex}
\usepackage{ifpdf}
\usepackage[T1]{fontenc}
\usepackage{times}
\usepackage{newtxmath}
\usepackage{textcomp}%
\usepackage{xcolor}%
\usepackage{hyperref}
\usepackage{url}
\usepackage{lipsum}
\newcommand{\la}{\lambda}

\articletype{METHODS ARTICLE}
\jname{Environmental Data Science}
\begin{document}

\begin{Frontmatter}

\title[XAI and Robustness]{Space-scale Exploration of the Poor Reliability of Deep Learning Models: the Case of the Remote Sensing of Rooftop Photovoltaic Systems}

\author[1,2]{Gabriel Kasmi}
\author[2,3]{Laurent Dubus}
\author[1]{Yves-Marie Saint-Drenan}
\author[1]{Philippe Blanc}

\authormark{Gabriel Kasmi \textit{et al}.}

\address[1]{\orgname{MINES Paris, Université PSL Centre Observation Impacts Energie (O.I.E.)}, \orgaddress{\city{Sophia-Antipolis}, \country{France}}}

\address[2]{\orgname{RTE France}, \orgaddress{\city{Paris}, \country{France}
}}

\address[3]{
\orgname{WEMC (World Energy \& Meteorology Council, UK)}
}

\authormark{Kasmi et al.}

%\received{31 January 2020}
%\revised{01 May 2020}
%\accepted{06 May 2020}


\abstract{Photovoltaic (PV) energy grows rapidly and is crucial for the decarbonization of electric systems. However, centralized registries recording the technical characteristifs of rooftop PV systems are often missing, making it difficult to accurately monitor this growth. The lack of monitoring could threaten the integration of PV energy into the grid. To avoid this situation, the remote sensing of rooftop PV systems using deep learning emerged as a promising solution. However, existing techniques are not reliable enough to be used by public authorities or transmission system operators (TSOs) to construct up-to-date statistics on the rooftop PV fleet. The lack of reliability comes from the fact that deep learning models are sensitive to distribution shifts. This work proposes a comprehensive evaluation of the effects of distribution shifts on the classification accuracy of deep learning models trained to detect rooftop PV panels on overhead imagery. We construct a benchmark to isolate the sources of distribution shift and introduce a novel methodology that leverages explainable artificial intelligence (XAI) and decomposition of the input image and model's decision in terms of scales to understand how distribution shifts affect deep learning models. Finally, based on our analysis, we introduce a data augmentation technique meant to improve the robustness of  deep learning classifiers to varying acquisition conditions. We show that our proposed approach outperforms competing methods. We discuss some practical recommendations for mapping PV systems using overhead imagery and deep learning models.} % 211 words

\policy{This paper analyzes the effects of distribution shifts on deep learning models trained to detect rooftop photovoltaic (PV) systems on aerial imagery by combining explainable artificial intelligence methods. It then proposes practical solutions grounded in this analysis to improve the robustness of these models, thus improving their reliability and facilitating the use of remote sensing techniques to improve insertion of rooftop PV systems into the grid.
}

\end{Frontmatter}

\section{Introduction}
\input{content/introduction}

\section{Related works}
\input{content/literature}

\section{Data}
\input{content/data}

\section{Methods}
\input{content/methods}

\section{Results}
\input{content/results}

\section{Discussion}
\input{content/conclusion}

\begin{appendix}\appheader
\input{content/appendix}
\end{appendix}
\newpage

\begin{Backmatter}

%\paragraph{Acknowledgments}
%We are grateful for the technical assistance of A. Author.

\paragraph{Funding Statement}
This research was supported by a grant from the ANRT (CIFRE funding 2020/0685) and was funded by the French transmission system operator RTE.

\paragraph{Competing Interests}
The authors declare no conflicts of interest.

\paragraph{Data Availability Statement}
Code for replicating the results of this paper can be found at: \url{https://github.com/gabrielkasmi/robust_pv_mapping}. Model weights can be found at: \url{https://zenodo.org/records/12179554}.

\paragraph{Ethical Standards}
The research meets all ethical guidelines, including adherence to the legal requirements of the study country.

\paragraph{Author Contributions}
Conceptualization, Gabriel Kasmi; Formal analysis, Gabriel Kasmi; Funding acquisition, Laurent Dubus; Investigation, Gabriel Kasmi; Methodology, Gabriel Kasmi; Project administration, Laurent Dubus; Software, Gabriel Kasmi; Supervision, Philippe Blanc, Yves-Marie Saint-Drenan and Laurent Dubus; Validation, Gabriel Kasmi; Writing – original draft, Gabriel Kasmi; Writing – review \& editing, Gabriel Kasmi, Philippe Blanc, Yves-Marie Saint-Drenan Laurent Dubus. All authors approved the final submitted draft

%\paragraph{Supplementary Material}
%State whether any supplementary material intended for publication has been provided with the submission.

\printbibliography

\end{Backmatter}

\end{document}

%% file: content/introduction.tex
% Context and need
Photovoltaic (PV) energy grows rapidly and is crucial for the decarbonization of electric systems \citep{haegel_terawatt-scale_2017}. The rapid growth of rooftop PV makes the estimation of the global PV installed capacity challenging as centralized data is often lacking \citep{kasmi_towards_2022,hu_what_2022}. Remote sensing of rooftop PV on orthoimagery with deep learning models is a blooming solution for mapping rooftop PV installations.  Deep learning-based pipelines became the standard method for remote sensing PV systems as works such as DeepSolar \citep{yu_deepsolar_2018} paved the way for country-wide mapping of PV systems using deep learning and overhead imagery. Recently, methods for mapping rooftop PV systems in many regions, especially in Europe, have been proposed \citep{mayer_deepsolar_2020,zech_predicting_2020,kausika_geoai_2021,lindahl_mapping_2023,frimane_identifying_2023,rausch_enriched_2020,kasmi_towards_2022}. Some of these works \citep{mayer_3d-pv-locator_2022,kasmi_towards_2022} introduced methods to estimate the technical characteristics of the PV systems (individual localization, orientation, PV installed capacity).

However, current approaches are sensitive to so-called distribution shifts \citep{de_jong_monitoring_2020}, i.e., differences between the training and testing data \citep{koh_wilds_2021}. This sensitivity limits their ability to generalize to new images without incurring significant accuracy drops, thus limiting their practical usability \citep{hu_what_2022} as the generated data lacks reliability \citep{de_jong_monitoring_2020}. Steps towards improving the reliability of the deep learning-based registries of rooftop PV systems have been taken, with \cite{hu_what_2022} and \cite{kasmi_towards_2022} discussing the practical evaluation of the mapping algorithms or \cite{li_understanding_2021} defining the minimum resolution to detect rooftop PV systems from overhead images. However, to date, only \cite{wang_poor_2017} studied the poor generalizability of PV mapping algorithms, but their evaluation framework was limited to the case study of two cities and one image dataset. Therefore, a comprehensive evaluation of the causes and consequences of distribution shifts on PV panel detectors and potential remedies to improve their robustness is missing despite its importance for improving their reliability in practice.

% Research gap
% Work is still needed to improve the robustness to acquisition condition of PV mapping algorithms and to propose a method to assess the reliability of the model's decisions.

% contributions
This work sets up an empirical benchmark using the training dataset BDAPPV\footnote{BDAPPV: Base de données d'apprentissage profond pour les installations photovoltaiques (Database for deep learning applied to PV systems).} \citep{kasmi_crowdsourced_2023} to disentangle the effects of distribution shifts on rooftop PV detectors. We then combine explainable artificial intelligence (XAI) methods to understand how these shifts affect the deep learning models. Based on our findings, we propose simple solutions that can be implemented to improve the robustness to distribution shifts of deep learning models trained to detect PV panels on overhead imagery. By improving the reliability and robustness of deep learning models for rooftop PV mapping, we aim to facilitate the mapping and, thus, the integration into the electric grid of rooftop PV. Code for replicating the results of this paper can be found at \url{https://github.com/gabrielkasmi/robust_pv_mapping} and model weights can be found at \url{https://zenodo.org/records/12179554}.

%% file: content/literature.tex
\subsection{Remote sensing of rooftop photovoltaic installations}

The remote sensing of rooftop PV systems is now a well-established field with early works dating back to \cite{malof_automatic_2015,malof_automatic_2016,golovko_development_2018,yuan_large-scale_2016}. The DeepSolar project \citep{yu_deepsolar_2018}  marked a significant milestone by mapping distributed and utility-scale installations over the continental United States using state-of-the-art deep learning models. Many works built on DeepSolar to map regions or countries, especially in Europe, covering areas such as North-Rhine Westphalia \citep{mayer_deepsolar_2020}, Switzerland \citep{casanova_instance-conditioned_2021}, Oldenburg in Germany \citep{zech_predicting_2020}, parts of Sweden \citep{lindahl_mapping_2023,frimane_identifying_2023}, Northern Italy \citep{arnaudo_comparative_2023}, the Netherlands \citep{kausika_geoai_2021} or the surroundings of Berkeley in California \citep{parhar_hyperionsolarnet_2021}. Several works even included GIS data to construct registries of PV installations \citep{kausika_geoai_2021,mayer_3d-pv-locator_2022,rausch_enriched_2020, kasmi_towards_2022}. In the current context of rapid rooftop PV growth \citep{haegel_terawatt-scale_2017,rte_france_energy_2022}, remote sensing of rooftop PV installations using deep learning and orthoimagery offers the potential to address the lack of systematic registration of small-scale PV installations \citep{kausika_gis4pv_2022,kasmi_towards_2022}.

However, current methods cannot be transposed from one region to another without incurring accuracy drops, thus limiting their practical usability \citep{hu_what_2022}. Unpredictable accuracy drops cast doubt on the reliability of the generated data \citep{de_jong_monitoring_2020}. Recently, \cite{kasmi_towards_2022} introduced a method aiming at indirectly assessing the accuracy of the detections by automatically comparing the registry generated with deep learning algorithms to reference data, which is often aggregated at the scale of the city. While this work enabled the quantification of the drop in accuracy encountered during deployment, no cues as to why the accuracy varied during deployment were discussed. In this work, we propose to study and mitigate the impact of these distribution shifts. 

\subsection{Sensitivity to distribution shifts}

\paragraph{Sensitivity to distribution shifts in remote sensing}

Distribution shifts, i.e., the sensitivity to the fact that "{\it the training distribution differs from the test distribution}" \citep{koh_wilds_2021} are ubiquitous in machine learning \citep{torralba_unbiased_2011}. The sensitivity to distribution shifts causes unpredictable performance drops, which can have dire consequences as models are deployed in safety-critical settings such as autonomous driving \citep{sun_shift_2022} or medical diagnoses \citep{pooch_can_2020}. 

In remote sensing applications, \cite{tuia_domain_2016} identified two primary sources of shifts in the input data models are sensitive to: variations in the geographical scenery, and the varying acquisition conditions. Following \cite{murray_zoom_2019}, we can add the ground sampling distance (GSD). The acquisition conditions encompass the conversion of a scene into a digital image and include all sources of variability in the input images caused by different sensors, exposure, attitude and altitude during acquisition, and atmospherical conditions. The ground sampling distance is the upper bound to the image resolution. The lower the ground sampling distance, the more detailed the image. In practice, the resolution is limited by the GSD and the image quality (noise, optical transfer function, intrinsic geometric consistency, etc). So far, the only work that investigated the poor reliability of deep learning systems applied to the remote sensing of PV panels is \cite{wang_poor_2017}. The authors argued that the generalization ability from one city to another depending on how "hard" to recognize the PV panels are. However, no proper definition of the "hardness" to recognize PV panels or a proper disentanglement the effect of each source of variability was carried out, and there was no prescription regarding model training or data preprocessing. 

\paragraph{Mitigating the sensitivity to distribution shifts} 

Numerous approaches have been introduced to mitigate the sensitivity to distribution shifts. We refer the reader to surveys such as \cite{zhou_domain_2023,tuia_domain_2016,guan_domain_2022,csurka_comprehensive_2017,csurka_unsupervised_2021} for reviews of these methods in various settings. One simple yet effective approach is to leverage data augmentations during training to incentivize the model to learn a given invariance during training. The aim is that the model is no longer sensitive to a given set of perturbations of the input images. Popular data augmentation methods consist in defining a method to generate as many perturbed samples as possible while preserving the semantic content of the image. To this end, AugMix \citep{hendrycks_augmix_2020} applies a random sequence with random weights of perturbations to the input image. Similarly, \cite{hendrycks_pixmix_2022} augment an input image with fractal patterns, and \cite{avidan_spectral_2022} perturb the Fourier spectrum of the input image. 

However, improving the robustness against distribution shifts is a long-tailed problem, meaning that unseen situations eventually arise, and all situations cannot be accounted for \citep{torralba_unbiased_2011,recht_imagenet_2019}. Therefore, we should first understand how distribution shifts affect a model's performance before implementing a mitigation method. To this end, we propose to leverage explainable artificial intelligence (XAI) methods. 

\subsection{Explainable artificial intelligence (XAI)}

Modern deep learning algorithms are often qualified as black boxes, meaning it is hard to fully grasp their inner workings. This black-box nature limits the applicability of machine learning in safety-critical settings \citep{achtibat_where_2022}. We can distinguish two main approaches for machine learning explainability: by-design interpretable models and post-hoc explainability \citep{parekh_cadre_2023}\footnote{\cite{flora_comparing_2022} note that there is no consensus yet in the literature regarding the use of the terms explainability and interpretability. Following \cite{flora_comparing_2022}, we say that a model is interpretable if it is {\it inhehirently} or {\it by design} interpretable and that a model is explainable if we can compute an {\it post-hoc} explanation of its decision.}. By-design interpretability aims at constructing models that are transparent and self-explanatory \citep{sudjianto_designing_2021}, e.g., the decision boundaries of a decision tree. On the other hand, post-hoc explainability seeks to explain a model's decision by highlighting important features contributing to this decision, without explicitly stating how these features affected the model. Methods such as class activation maps (CAMs, \cite{zhang_understanding_2017}), which plot a heatmap of the important image regions for the classification of this image, fall into this category.

\paragraph{XAI methods for model debugging} 

One of the main motivations for XAI is to inspect the decision of models to assess whether they relied on relevant factors to make predictions. Several works highlighted biases in the decision process, such as the reliance on spurious features. \cite{lapuschkin_unmasking_2019} leveraged the GradCAM \citep{selvaraju_grad-cam_2017} to show how classifiers could rely on watermarks rather than relevant areas of the input image for horses classification, thus highlighting a so-called "Clever Hans" \citep{pfungst_clever_1911} effect\footnote{Clever Hans was a horse that appeared to perform arithmetic and other intellectual tasks but was actually responding to subtle cues from his handler. In machine learning, it is used as an example of the reliance of deep learning models on spurious features (e.g., background) rather than causal features (e.g., shape of an object) for tasks such as classification. See \cite{lapuschkin_unmasking_2019} for more details.}. CAMs \citep{zhang_understanding_2017} have also been used to understand the behavior of convolutional neural networks (CNNs) in medical imagery classification by \cite{zhang_grad-cam_2021}. Another example of usage of XAI tools to understand and debug a model was proposed by \cite{dardouillet_explainability_2023}, who leveraged SHapley Additive exPlanations (SHAP, \cite{lundberg_unified_2017}) to understand a model deployed for oil slick pollution detection on the sea surface. In this work, we go one step further and show how combining post-hoc and by-design interpretable XAI methods can help understand and mitigate the sensitivity to distribution shifts of CNNs deployed for mapping PV systems from overhead imagery. 

%% file: content/data.tex
\begin{figure}[h]
\centering
\includegraphics[width=0.7\textwidth]{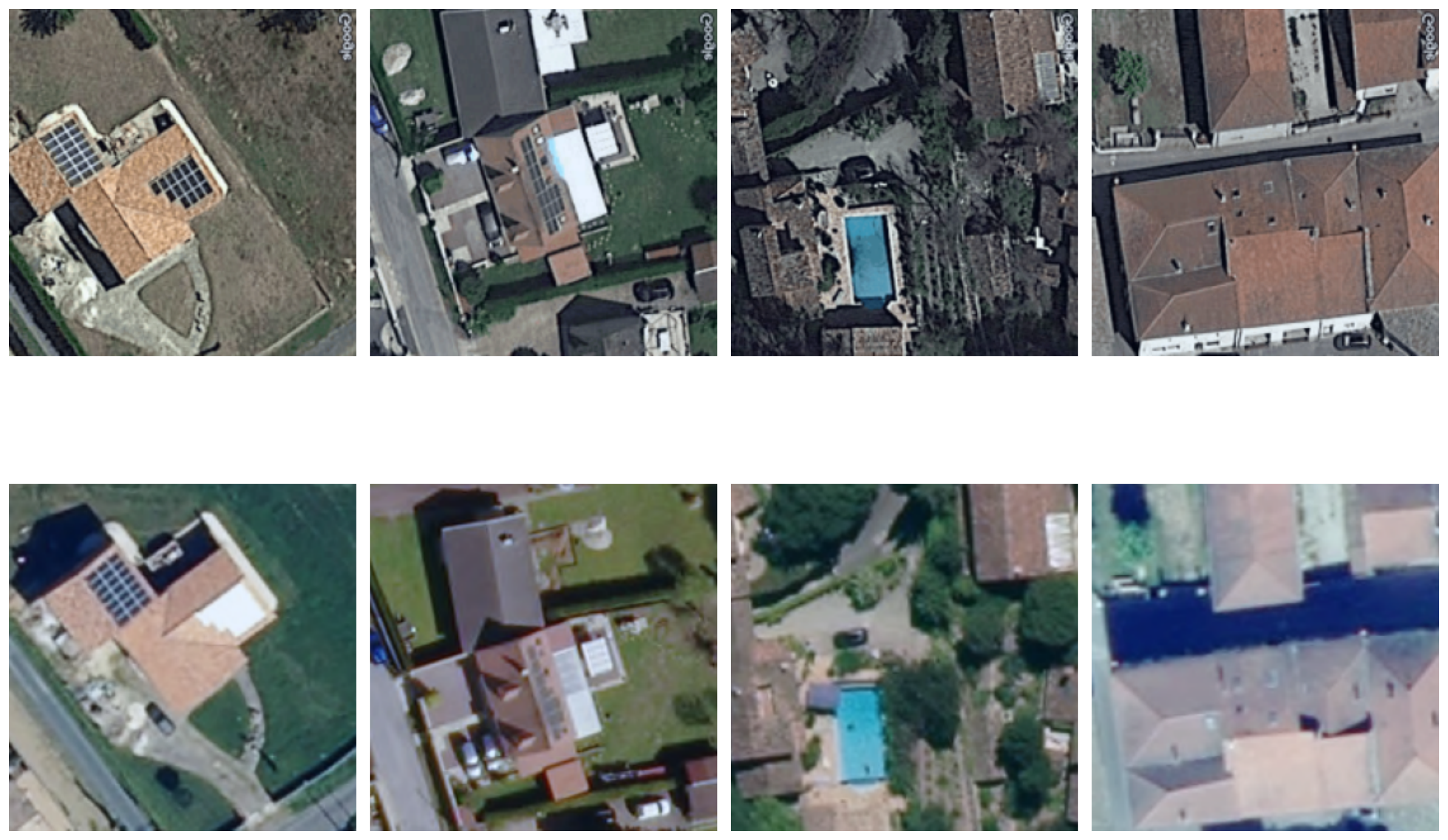}
\caption{Examples of images of the same PV panels but with different providers and acquisition dates (Up Google, down: IGN)}\label{fig:examples}
\end{figure}

We consider the crowdsourced training dataset BDAPPV\footnote{The dataset is accessible here: \url{https://zenodo.org/records/7358126}.} \citep{kasmi_crowdsourced_2023}. This dataset contains annotated images of 28,000 PV panels in France and neighboring countries. This dataset also proposes annotations of images that depict the same PV panels, but from two different image providers: images coming from the Google Earth Engine (hereafter referred to as "Google," \cite{gorelick_google_2017}) and from the IGN \citep{ign_bd_2024}, the French public operator for geographic information. We have double annotations for 7,686 PV systems. It allows us to assess the impact of the acquisition conditions as the only change factor between two images is the varying acquisition condition: the semantic content (the PV panel and its surroundings) remains (almost) unchanged. The native ground sampling distance (GSD) of Google images is 10 cm/pixel and 20 cm/pixel for IGN images. We define the acquisition conditions as the properties of the technical infrastructure (airborne or spaceborne, camera type, image quantization, and postprocessing) and the atmospheric and meteorological conditions the day the image was taken. \autoref{fig:examples} plots examples of images coming from BDAPPV. These images depict the same scene for both providers (although potentially at different dates).

%% file: content/methods.tex
We aim to explain why convolutional neural networks (CNNs) applied to detect PV panels on orthoimages are sensitive to distribution shifts. We first construct a benchmark to isolate the effect of the three main sources of distribution shifts on overhead images highlighted by \cite{tuia_domain_2016} and \cite{murray_zoom_2019}. On our benchmark dataset, we then compare the behavior of a by-design interpretable model, the Scattering transform \citep{bruna_invariant_2013}, with the predictions of our CNN to understand on which scales the CNN relies and how it is affected by a disruption of these scales. To verify that our mechanism is correct, we leverage the wavelet scale attribution method (WCAM, \cite{kasmi_assessment_2023}), which is a post-hoc explainability method, to isolate the important scales in the predictions of our black-box CNN model. We chose the Scattering transform and the WCAM because these methods rely on a decomposition of the input image in the space-scale (wavelet domain), which is particularly well suited in the case of remote sensing images since the scales, expressed in pixels on images, are indexed in meters and can thus point towards actual elements depicted on the images\footnote{In appendix \ref{sec:limitations}, we provide further evidence of the limitation of "traditional" feature attribution methods for explaining the false detection of deep learning models in our use case.}. Finally, based on our findings, we propose a data augmentation method to improve the robustness of CNNs and draw some lessons regarding the choice of image data. 

\subsection{Disentangling the sources of distribution shifts on overhead images}

BDAPPV features images of the same installations from two providers and records the approximate location of the PV installations. Using this information, we can define three test cases to disentangle the distribution shifts that occur with remote sensing data: the resolution, the acquisition conditions, and the geographical variability. We train a ResNet-50 model \citep{he_deep_2016} on Google images downsampled at 20 cm/pixel of resolution and evaluate it on three datasets: a dataset with Google images at their native 10 cm/pixel resolution ("Google 10 cm/pixel"), the IGN images with a native 20 cm/pixel resolution ("IGN") and Google images downsampled at 20 cm/pixel located outside of France ("Google OOD\footnote{OOD: out-of-distribution.}"). We add the test set to record the test accuracy without distribution shift ("Google baseline"). We only do random crops, rotations, and ImageNet normalizations during training. 

\begin{figure}[h]
\centering
\includegraphics[width=0.7\textwidth]{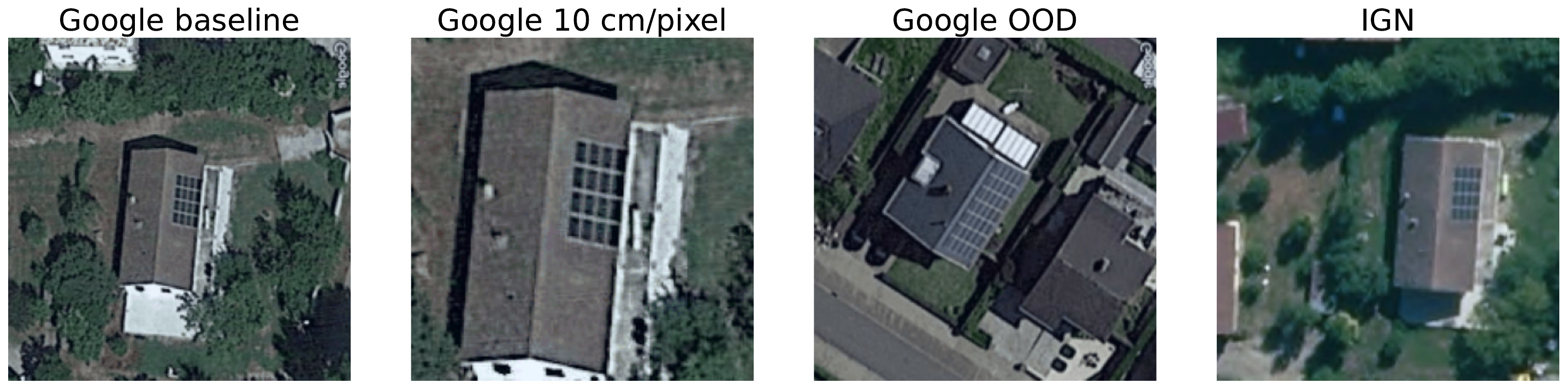}
\caption{Test images on which a model trained on Google images (downsampled to 20 cm/px of GSD, "Google baseline") is evaluated. "Google 10 cm/pixel" corresponds to the source Google image, before downsampling and evaluates the effect of varying ground sampling distances. "Google OOD" corresponds to Google images taken outside of France. "IGN" corresponds to images depicting the same installations as Google baseline but with a different provider}\label{fig:test-images}
\end{figure}

\autoref{fig:test-images} plots examples of the different test images to disentangle the effects of distribution shifts. The baseline and IGN images represent the same panel at the same spatial resolution. The Google 10 cm/pixel depicts the same scene but with the native resolution of Google images. Finally, the OOD test set contains images located outside of France.

\subsection{Space-scale decomposition of a model's decision process}

\subsubsection{Background: the wavelet transform of an image}

\begin{figure}[h]
    \centering
    \includegraphics[width=.8\textwidth]{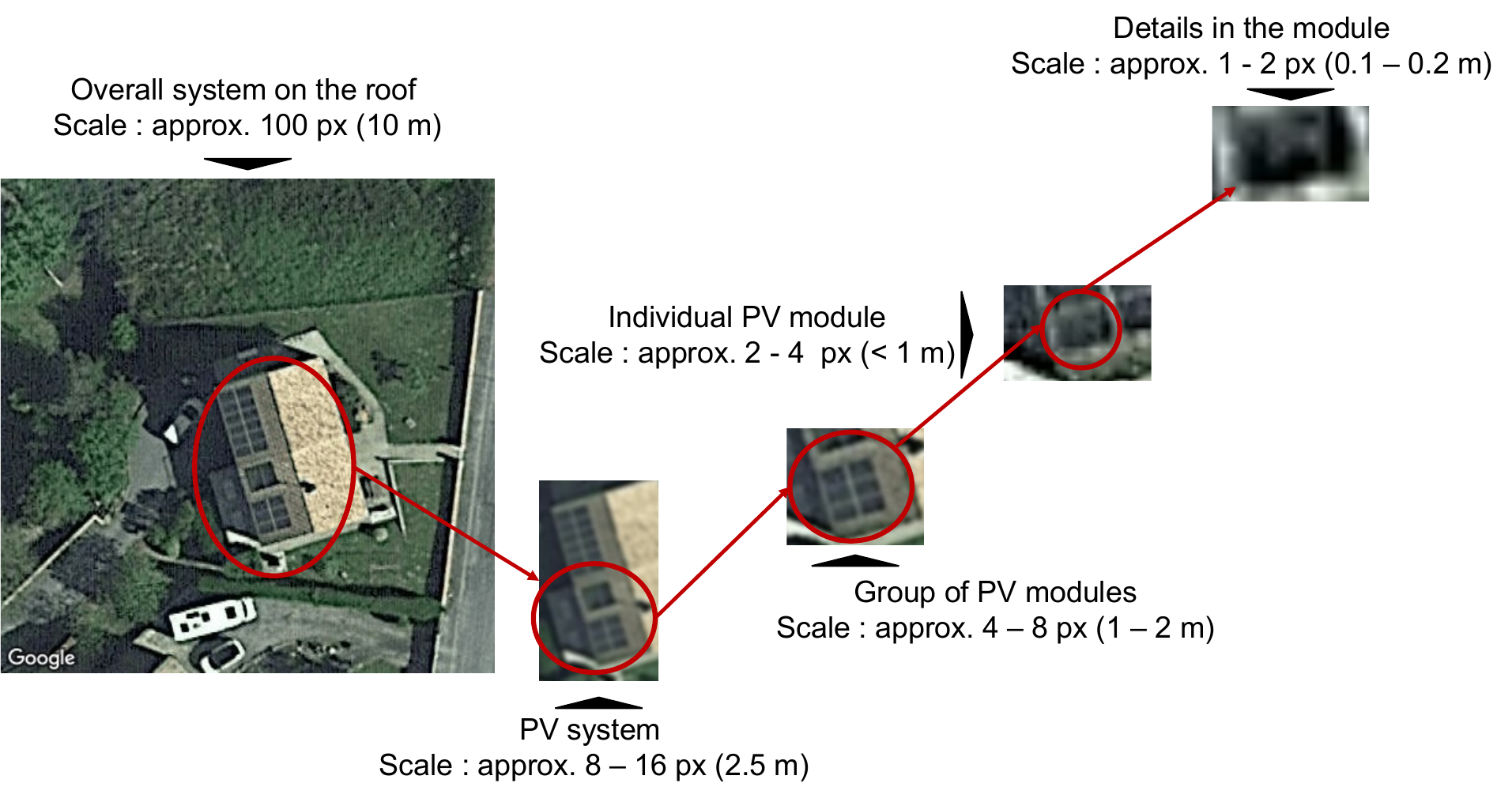}
    \caption{Decomposition of a PV panel into scales}
    \label{fig:pv-scales}
\end{figure}

\paragraph{Motivation and definition} We propose to analyze the decision process of an off-the-shelf CNN model through the lenses of the space-scale or wavelet decomposition. Wavelets are a natural tool to decompose an image into scales while maintaining local analysis in space: they provide a single space-scale decomposition. As scales are indexed in terms of actual distances on the ground, we can directly identify the important objects that contribute to a model's decision by studying the important scales. \autoref{fig:pv-scales} illustrates the objects that can be found at different scales of an orthoimage. 

A wavelet is an integrable function $\psi\in L^2 (\mathbb{R})$ with zero average, normalized and centered around 0. Unlike a sinewave, a wavelet is localized in space and in the Fourier domain. It implies that dilatations of this wavelet enable to scrutinize different scales while translations enable to scrutinize spatial location. In other words, scales correspond to different spatial frequency ranges or spectral domains. %(see \autoref{fig:dyadic-correspondance} for an illustration). 

%\begin{figure}[h]
    %\centering
    %\includegraphics[width=\textwidth]{figures/decomposition.pdf}
    %\caption{Correspondence between the scales in the wavelet domain and the frequency ranges in the Fourier domain. In this diagram, $f_s$ corresponds to the highest frequency contained in the signal. Inspired by \cite{chen_centralized_2019}.}
    %\label{fig:dyadic-correspondance}
%\end{figure}

To compute an image's (continuous) wavelet transform (CWT), one first defines a filter bank $ \mathcal{D}$ from the original wavelet $\psi$ with the scale factor $s$ and the 2D translation in space $u$. We have
\begin{equation}
        \mathcal{D} = \left\{
        \psi_{s,u}(x) = \frac{1}{\sqrt{s}}\psi\left(\frac{x-u}{s}\right)
    \right\}_{u\in\mathbb{R}^2,\;s\ge 0},
\end{equation}
where $\vert \mathcal{D}\vert =J$, and $J$ denotes the number of levels. The computation of the wavelet transform of a function $f\in L^2(\mathbb{R})$ at location $x$ and scale $s$ is given by
\begin{equation}\label{eq:wt}
    \mathcal{W}(f)(x,s) = \int_{-\infty}^{+\infty}
    f(u) \frac{1}{\sqrt{s}} \psi^*\left(\frac{x-u}{s}\right)
    \mathrm{d}u,
\end{equation}
which can be rewritten as a convolution \citep{mallat_wavelet_1999}. Computing the multi-level decomposition of $f$ requires applying \autoref{eq:wt} $J$ times with all dilated and translated wavelets of $\mathcal{D}$. \cite{mallat_theory_1989} showed that one could implement the multi-level dyadic decomposition of the discrete wavelet transform (DWT) by applying a high-pass filter $H$ to the original signal $f$ and subsampling by a factor of two to obtain the {\it detail} coefficients and applying a low-pass filter $G$ and subsampling by a factor of two to obtain the {\it approximation} coefficients. Iterating on the approximation coefficients yields a multi-level transform where the $j^{th}$ level extracts information at resolutions between $2^j$ and $2^{j-1}$ pixels. The detail coefficients can be decomposed into various rotations (usually horizontal, vertical, and diagonal) when dealing with 2D signals (e.g., images). %\autoref{fig:dyadic-correspondance} illustrates the multi-level decomposition of a signal into approximation and detail coefficients. The detail coefficients at level $k$ contain the frequencies located between $\displaystyle{
%\frac{1}{2^{k+1}}
%}\pi$ and $\displaystyle{
%\frac{1}{2^{k}}
%}\pi$ where $\pi = \displaystyle{
%\frac{f}{f_s}
%}$ and $f_s$ is the highest frequency in the signal. 

\begin{figure}[h]
    \centering
    \includegraphics[width = .9\textwidth]{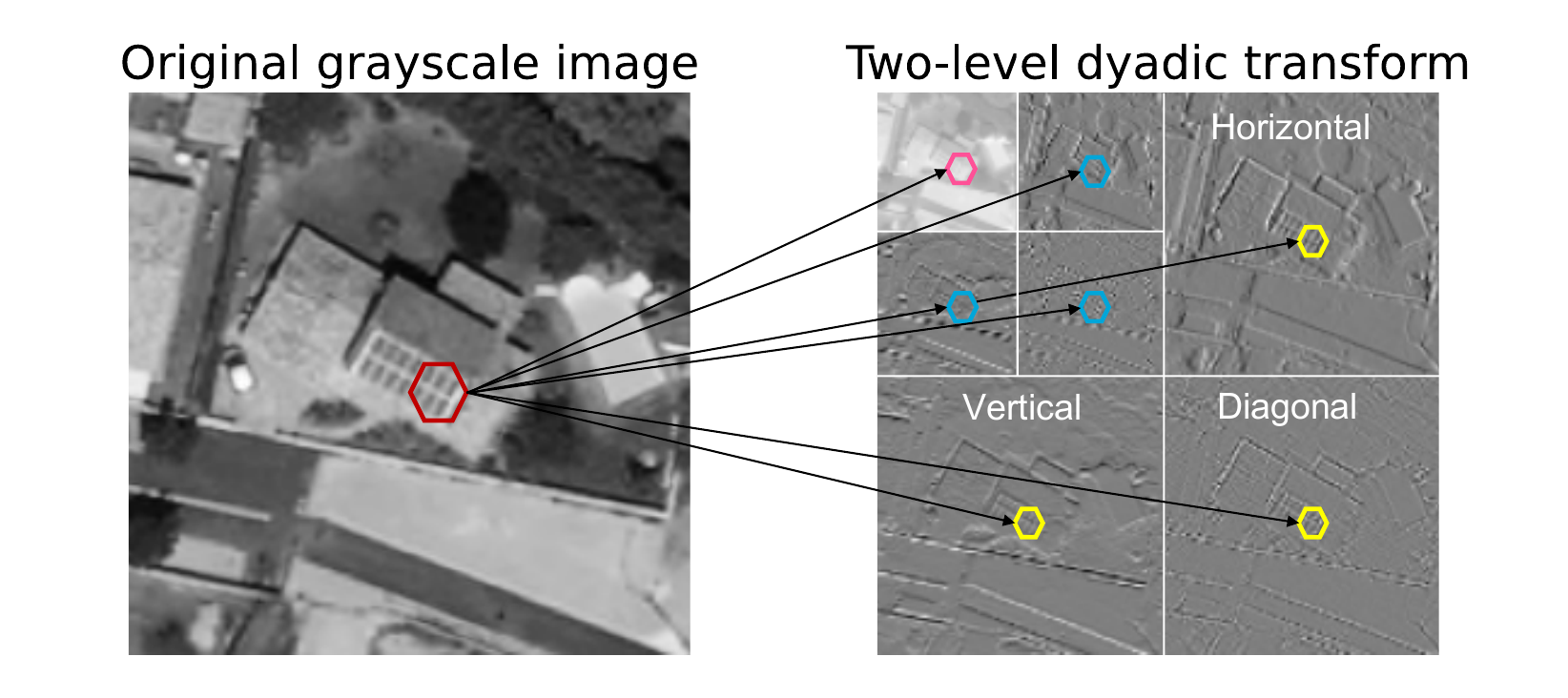}
    \caption{Image and associated two-level dyadic wavelet transform with indications to interpret the wavelet transform of the image. "Horizontal," "diagonal," and "vertical" indicate the direction of the detail coefficients. The direction is the same at all levels}
    \label{fig:reading-wavelet}
\end{figure}

\paragraph{Interpreting the wavelet transform of an image} \autoref{fig:reading-wavelet} illustrates how to interpret the (two-level) wavelet transform of an image. Reading is the same for any multi-level decomposition. The right image plots the two-level dyadic decomposition of the original image depicted on the left. Following this transform, the localization on the image highlighted by the red polygon can be decomposed into six detail components (marked yellow and blue) and one approximation component (marked pink). The yellow components correspond to the details at the 1-2 pixel scale, and the blue components to the details at the 2-4 pixel scale. For each location, the wavelet transform summarizes the information contained in the image at this scale and location.

\subsubsection{By design interpretable XAI method: the Scattering transform}

\begin{figure*}[h]
\centering
\includegraphics[width=\textwidth]{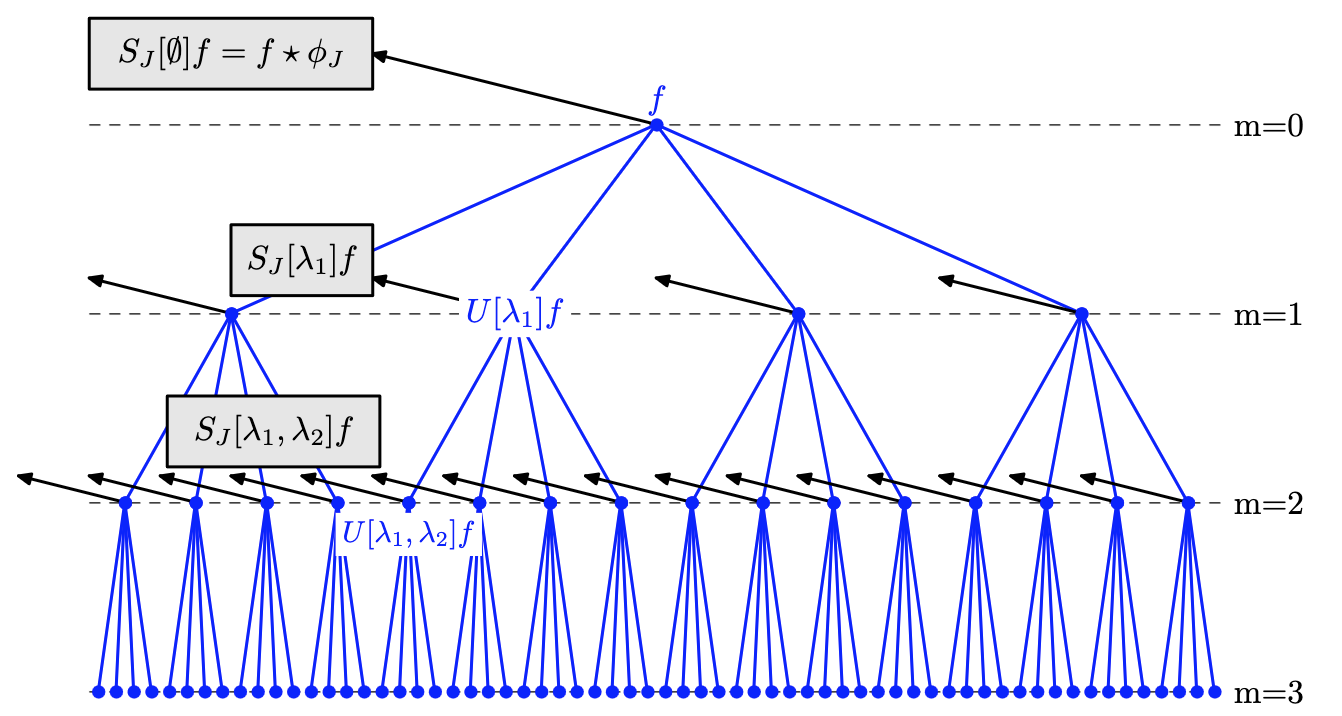}
	\caption{A scattering propagator $U_J$ applied to $x$ 
computes each
$U[\la_1] x = |x \star \psi_{\la_1}|$ and outputs 
$S_J [\emptyset] x = x \star \phi_{2^J}$ (black arrow).
Applying $U_J$ to each $U[\la_1] x$ computes all
$U[\la_1,\la_2]x$ and outputs 
$S_J[\la_1] = U[\la_1] \star \phi_{2^J}$ (black arrows).
Applying  
$U_J$ iteratively to each $U[p]x$ outputs $S_J[p]x = U[p]x \star \phi_{2^J}$
(black arrows) and computes the next path layer.
Figure borrowed from \cite{bruna_invariant_2013}. Note: On the image, the input $x$ corresponds to $f$ and $\lambda=2^j r$ is a frequency variable corresponding the the $j^\textrm{th}$ scale with $r$ rotations}
	\label{fig:scattering-cascade}
\end{figure*}

The Scattering transform \citep{bruna_invariant_2013} is a deterministic feature extractor. CNNs and the Scattering transform share the same multi-level architecture, where the previous layer's output is passed onto the next after a nonlinearity is applied. The nonlinearities in a CNN are generally rectified linear units (ReLU), whereas in the Scattering transform, it is a modulus operation. Unlike CNNs, whose kernel coefficients are learned during training, the coefficients of the Scattering transform are fixed. \cite{bruna_invariant_2013} showed that the Scattering transform computes representations from an input image that share the same properties of translational invariance as the representations computed with a CNN. The advantage of the Scattering transform is that as filters are fixed, we can know precisely what information they extract from the input image. \autoref{fig:scattering-cascade} summarizes the feature extraction process of the Scattering transform.  

The input image $x$ is downsampled, and a wavelet filter $\phi$ is applied in $J$ directions. The wavelet coefficients at that scale are retrieved (black arrows), and the image is passed onto the next layer (blue arrows). As the depth increases, the spatial extent covered by the filters decreases. At each spatial location, one takes the modulus of the wavelet transform to compute a scale invariant representation which indicates the amount of "energy" in the image at this scale and localization. 

The Scattering transform is parameterized by the number $m$ of layers and the number $J$ of orientations. We have a total of $mJ + m^2 J(J-1)/2$ coefficients. At the end of the decomposition, the features, i.e., the scattering coefficients, are flattened into a single vector of size $mJ + m^2 J(J-1)/2$. We can identify to which scale, location, and orientation on the input image this feature corresponds. 

We implement three variants of the Scattering transform with depths $m$ varying from one to three levels. \cite{bruna_invariant_2013} stated that first-order coefficients were insufficient to discriminate between two very different images but that coefficients of order $m=2$ could. We consider $J=8$ orientations. We stack the scattering coefficients into a vector of dimension $mJ + m^2 J(J-1)/2$, akin to the penultimate layer of a CNN. We train a linear classifier on this feature vector. Our implementation of the Scattering transform is based on the Python library Kymatio \citep{andreux_kymatio_2020}.

\subsubsection{Post-hoc XAI method: the Wavelet scale attribution method (WCAM)}

Traditional feature attribution methods  \citep{simonyan_very_2015,selvaraju_grad-cam_2020,petsiuk_rise_2018} highlight the important areas for the prediction of a classifier in the pixel (spatial) domain. The WCAM \citep{kasmi_assessment_2023} generalizes attribution to the wavelet (space-scale domain). The WCAM provides us with two pieces of information: where the model sees and what scale it sees at this location. The decomposition of the prediction in terms of scales points towards actual elements on the input image since on orthoimagery scales are indexed in meters. For example, on Google images, details at the 1-2 pixel scale correspond to physical objects with a size between 0.1 and 0.2 m on the ground. Thus, we know what the model sees as a panel; we can interpret it and assess whether it is sensitive to varying acquisition conditions. We refer the reader to appendix \ref{sec:wcam-computation} or to \cite{kasmi_assessment_2023} for more details on the computation of the WCAM. 

\paragraph{Reading a WCAM}

\begin{figure}[h]
    \centering
    \includegraphics[width=.6\textwidth]{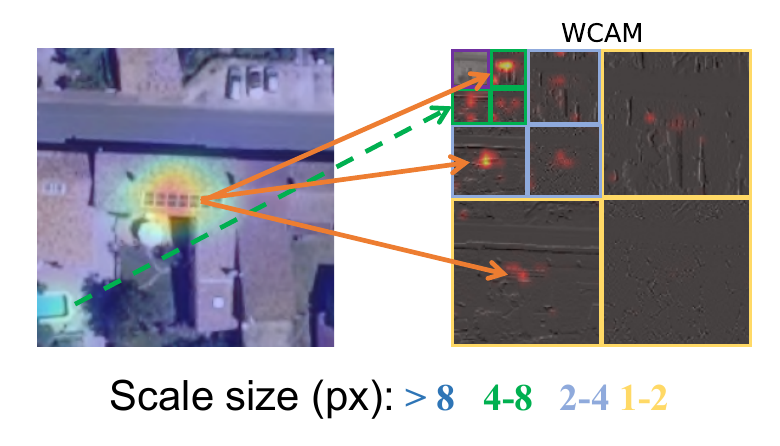}
    \caption{Decomposition in the wavelet domain of the important regions for a model's prediction with the WCAM}
    \label{fig:wcam-reading}
\end{figure}

\autoref{fig:wcam-reading} presents an example of an explanation computed using the WCAM. On the right panel, we can see the important areas in the model prediction highlighted in the wavelet domain. On the left panel, we can see the spatial localization of the important components. We can see two main spatial locations: the center of the image, which depicts the PV panel, and the bottom left, which depicts a pool. Disentangling the scales, we can see that the PV panel's importance spreads across three scales (orange arrows), while the pool is only important at the 4-8 pixel scale. This underlines that the model focuses on the PV panel because it sees details ranging from small details in the PV modules to the cluster of modules.  

\subsection{Improving the robustness through data augmentations}

\paragraph{Improving the robustness to noise and scale distorsions} Since we know that varying acquisition conditions induce perturbations \citep{lone_noise_2018}, which primarily affects high-frequency components (i.e., the finest scales), we introduce two data augmentation techniques that aim at reducing the reliance on high-frequency components: Gaussian blurring ("Blurring") and Blurring + wavelet perturbation (WP). Blurring consists of a fixed image blur, while the Blurring + WP also perturbs the wavelet coefficients of the image at specific scales to force the model to learn potential distortions incurring at these scales. We refer the reader to the appendix \ref{sec:augmentation-plot} for more details on the data augmentation strategies and a review of the hyperparameters. \autoref{fig:augmentations} in appendix \ref{sec:plot-augmentation} illustrates the effect of the different data augmentation techniques. We compare this approach with a baseline without augmentations (labeled "ERM" for empirical risk minimization, \cite{vapnik_nature_1999}) and existing data augmentation techniques. We also include an "Oracle", i.e. an ERM trained on IGN images. The Oracle is a proxy for the best possible performance on the domain of interest.

\paragraph{Comparisons with existing approaches} The literature on robustness to image corruptions \citep{hendrycks_benchmarking_2019} proposed numerous data augmentation methods to improve the robustness of classification models to image corruptions \citep{hendrycks_augmix_2020,hendrycks_pixmix_2022,cubuk_randaugment_2020,cubuk_autoaugment_2019,geirhos_imagenet-trained_2019}. We consider the AugMix method \citep{hendrycks_augmix_2020} and the recently-proposed RandAugment \citep{cubuk_randaugment_2020} and AutoAugment \citep{cubuk_autoaugment_2019} methods. These methods apply a random composition of perturbations to images during training to learn an invariance against these perturbations. We do not consider the case of training from multiple sources as our setting is that we wish to generalize to unseen images (either temporally or spatially, so we cannot incorporate knowledge about these images). For comparison, we include an Oracle, a model trained on the target dataset (i.e., a model trained on IGN images). The Oracle is a proxy for the best performance that can be achieved on this dataset.

%% file: content/results.tex
\subsection{Deep models are mostly sensitive to varying acquisition conditions, leading to an increase in the number of false negatives}

\autoref{tab:baseline-results} shows the results of the decomposition of the effect of distribution shifts into three components: resolution, acquisition conditions, and geographical shift. We can see that the F1 score drops the most when the model faces new acquisition conditions. The second most significant impact comes from the change in the ground sampling distance, but the performance drop remains relatively small compared to the effect of the acquisition conditions (which can also be assimilated to variations in the image quality). In our framework, there is no evidence of an effect of the geographical variability once we isolate the effects of the acquisition conditions and ground sampling distance. This effect is probably underestimated, as images of our dataset that are not in France are near France. However, the effect of the acquisition conditions is sizeable enough to seek methods for addressing it.

\begin{table}[h]
\small
  \centering
  \vspace{0mm}\caption{\textbf{F1 Score} and decomposition in true positives, true negatives, false positives, and false negatives rates of the classification accuracy of a CNN model trained on Google images ({\it Google baseline}) and tested on the three instances of distributions shifts: GSD ({\it Google 10 cm/px}), the geographical variability ({\it Google OOD}) and the acquisition conditions ({\it IGN}). Values highlighted in red indicate the worst performance and values in orange the second-to-last worse performance.}\label{tab:baseline-results} 
  \resizebox{\textwidth}{!}{\begin{tabular}{r c c c c c}
  \toprule
   \multicolumn{1}{c}{Model}& F1 Score ($\uparrow$) & True positive rate ($\uparrow$) & True negative rate ($\uparrow$) & False positive rate ($\downarrow$) & False negative rate ($\downarrow$) \\
   \midrule
    Google baseline &  0.98  & 0.99 & 0.98 & 0.02& 0.01 \\ 
    Google 10 cm/px & {\color{orange}0.89} &  0.81 & 1.00 & 0.00& {\color{orange}0.19}\\
    Google OOD &0.98 &  0.99 & 0.98 & 0.02& 0.01\\
    IGN & {\color{red}0.46} & 0.32 & 0.95 & 0.03& {\color{red}0.68}\\

  \bottomrule
  \end{tabular}}
  \end{table}

% Résultat du benchmark

\subsection{The Scattering transform shows that clean, fine-scale features are transferable but poorly discriminative}

\paragraph{Discriminative and transferable features} In the following, we distinguish between two kinds of features: the {\it discriminative} and the {\it transferable} features. Discriminative features enable the model to discriminate well between PV and non-PV images. Relying on discriminative features ensures a low number of false positives. On the other hand, transferable features correspond to features that generalize well across domains. If a model relies on transferable features, its performance should remain even across domains. Ideally, we would like a model to rely on discriminative and transferable features. Analysis of the errors of the CNN using the Scattering transform highlights a potential trade-off between transferable and discriminative features.

\paragraph{Accuracy of the Scattering transform} \autoref{tab:results-scattering} presents the accuracy results of the Scattering transform and compares it with a random classifier and the ERM (which is the same model as the one evaluated in \autoref{tab:baseline-results}). We can see that the performance on the source domain lags behind the performance of the CNN, but the Scattering transform generalizes better to IGN than the CNN. However, this comes at the cost of a high false positive rate. \autoref{tab:results-scattering-full} in appendix \ref{sec:appendix-stm} presents similar accuracy results for variants of the Scattering transform model in the depth and number of features.

  \begin{table}[h]
\small
  \centering
  \setlength{\tabcolsep}{2pt}
  \vspace{0mm}\caption{\textbf{F1 Score} and decomposition in true positives, true negatives, false positives and false negative rate of the classification accuracy of the Scattering Transform model trained on Google images and deployed on IGN images. Best results are {\bf bolded}, values in red highlight problematic cases.}\label{tab:results-scattering}
  \resizebox{\textwidth}{!}{\begin{tabular}{c r c c c c c}
  \toprule
   \multicolumn{1}{c}{Model}&Dataset  & F1 Score ($\uparrow$) & True positive rate ($\uparrow$) & True negative rate ($\uparrow$) & False positive rate ($\downarrow$) & False negative rate ($\downarrow$) \\
    \midrule
    \multirow{2}{*}{Scattering transform}&{\it Google baseline}&0.57&0,89	&0,10	&\color{red}{0,56}	&0,48\\
                        &IGN&{\bf 0.59}&0,54	&0,31	&\color{red}{0,62}	&0,54\\
                        \cmidrule{2-7}
    \multirow{2}{*}{CNN (ERM)}& {\it Google baseline} &  {\bf 0.98}  & 0.99 & 0.98 & 0.02& 0.01 \\ 
    &{\it IGN} & 0.46 & 0.32 & 0.95 & 0.03& 0.68\\
    \cmidrule{2-7}
    \multirow{2}{*}{Random classifier}&{\it Google baseline}&0.47 & 0,5&	0,50&	0,55&	0,45\\
    &{\it IGN}&0.47 & 0,50&	0,50&	0,56&	0,44\\
  \bottomrule
  \end{tabular}}
  \end{table}

\paragraph{Implications for the CNN} We know which features the Scattering transform relies on. It leverages information at the two-pixel scale after downsampling the input image. In other words, the Scattering transform makes predictions based on {\it clean} features at the two-pixel scale. Therefore, we can deduce that these features are {\it transferable}, as the performance remains even across datasets, but not very {\it discriminative} as the false positives rate is high (across both domains). 

On the other hand, the CNN should rely on {\it discriminative} features, which are located at coarser scales than 8 pixels, and on noisy features. In section \ref{sec:distorsion}, we investigate how the distortion of the input image's coarse scales impacts the CNN's decision process and the shift in its predicted probability. In section \ref{sec:data-aug}, we discuss how noise in input images affects the generalization ability of the CNN.

\subsection{CNNs are sensitive to the distorsion of coarse-scale discriminative features}\label{sec:distorsion}

\paragraph{Predicted probability shifts}

The CNN outputs a predicted probability of a PV panel on the input image. When evaluating the CNN on the same scene coming from two providers, we compute {\it predicted probability shift} $\Delta p = \vert p_{ign} - p_{google}\vert$ when the model trained on Google is evaluated on IGN images. By construction, $\Delta p\in[0,1]$. If $\Delta p=0$, the predicted probability did not change when changing the provider. On the other hand, if $\Delta p\to 1$, then it means that the model made a different prediction solely because of the new acquisition condition. 

\paragraph{Correlations between the probability shift and low-scale similarity of the images} 

For all images in our test set ($n=4321$), we compute the similarity between the approximation coefficients at the 3rd level (corresponding to details above 8 pixels) and the predicted probability shift of the model. We expect that the greater the dissimilarity, the higher the shift in probability. We evaluate the dissimilarity between the low-scale details using the Structural similarity index measure (SSIM, \cite{wang_image_2004}) and the Euclidean distance between the images. The SSIM takes values between -1 and 1, where 1 indicates perfect similarity, 0 indicates no similarity, and -1 indicates perfect anti-correlation. On the other hand, the Euclidean distance takes positive values; the greater the distance, the greater the dissimilarity between the images. 

As expected, we obtain a negative Pearson correlation coefficient equal to -0.41 (with a $p$-value $<10^{-5}$) between the input images' SSIMs and the predicted probability shift. Using the Euclidean distance, we obtain a correlation coefficient of 0.250 ($p<10^{-5}$). These results back the idea that the CNN is sensitive to low-scale perturbations of the input image, which results in a shift in the predicted probability. 

\paragraph{Visualization of the model's response with the WCAM}

\begin{figure}[h]
    \centering
    \includegraphics[width=.7\textwidth]{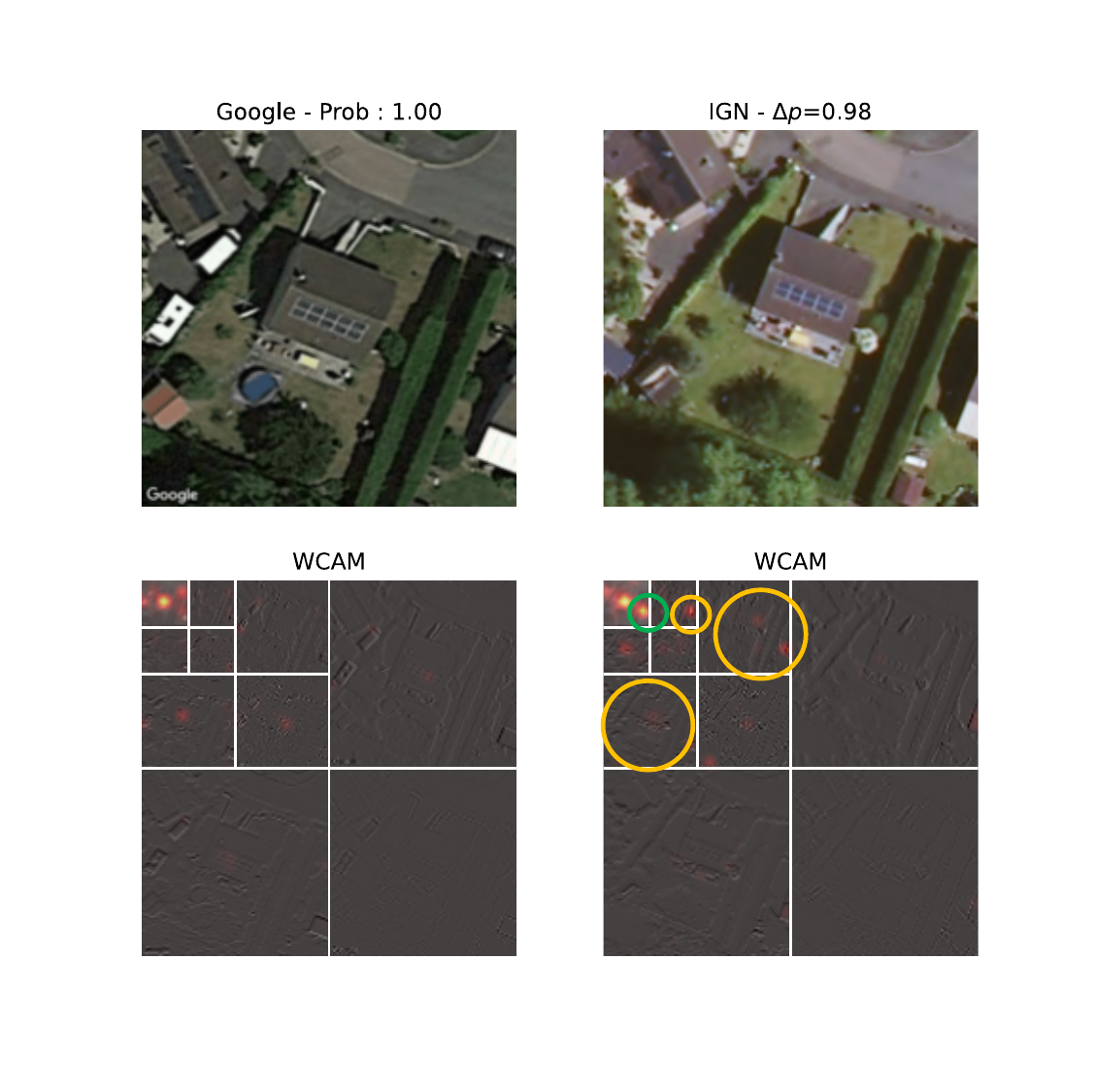}
    \caption{Analysis with the WCAM of the CNNs prediction on an image no longer recognized as a PV panel}
    \label{fig:wcam-example}
\end{figure}

The WCAM disentangles the important scales in a model's prediction. It enables us to see which scales were disrupted. On \autoref{fig:wcam-example}, we present an example of an image that was initially identified as a PV panel but turned out to be no longer recognized. 

We can see that in both cases, the approximation details are important in the model's prediction. The model responds to distortions at this scale by no longer focusing on a single area. Indeed, the model weights more components located at the 2-4 and 4-8 pixel scale (orange circles), which were not as important at the beginning. At the level of the perturbed scales, we can also witness that the model is disrupted by factors lying next to the PV panel (green circle). We supply more examples of such cases in the appendix \ref{sec:additional-figues} and discuss quantitative analysis of this result in appendix \ref{sec:scale-embeddings}.  

\subsection{Pathways towards improving the robustness to acquisition conditions}

%% méthode proposée et résultats

\subsubsection{Blurring and wavelet perturbation improve accuracy}\label{sec:data-aug}

\autoref{tab:mitigation} reports the results of our data augmentation techniques and compares them with existing methods. We can see that augmentations that explicitly discard small scales (high frequencies) information perform the best%\footnote{This result is further underlined by the fact that the Oracle -- trained on IGN -- evaluated on Google performs better than the model trained on Google images and tested on IGN. As IGN images have less information in the highest frequencies, lowering the reliance on the highest frequencies is essential to guarantee a good generalization to new acquisition conditions.}
. However, the blurring method sacrifices the recall (which drops to 0.6) to improve the F1 score. On \autoref{tab:mitigation}, this can be seen by the increase in false positives. Therefore, this method is unreliable for improving the robustness to acquisition conditions. 

On the other hand, adding wavelet perturbation (WP) improves the accuracy of the classification model without sacrificing the precision or the recall. While the drop in accuracy is still sizeable compared to the Oracle, the gain is consistent compared to other data augmentation techniques. Compared to RandAugment, the best-benchmarked method, our Blurring + WP is closer to the targets regarding true positives and true negatives and makes lower false negatives. This experiment shows that it is possible to consistently and reliably improve the robustness of acquisition conditions using a data augmentation technique, which does not leverage any information on the IGN dataset. 

  \begin{table}[h]
\small
  \centering
  \vspace{0mm}\caption{\textbf{F1 Score} and decomposition in true positives, true negatives, false positives, and false negatives rate for models trained on Google with different mitigation strategies. Evaluation on IGN images. The oracle corresponds to a model trained on IGN images with standard augmentations. The best results are {\bf bolded} and second best \underline{underlined}. Values in red highlight problematic cases.}\label{tab:mitigation}
  \resizebox{\textwidth}{!}{\begin{tabular}{r c c c c c}
  \toprule
   
   \multicolumn{1}{c}{Model}& F1 Score ($\uparrow$) & True positive rate ($\uparrow$) & True negative rate ($\uparrow$) & False positive rate ($\downarrow$) & False negative rate ($\downarrow$) \\
    \midrule
    Oracle & 0.88 & 0.96 & 0.82 & 0.18 & 0.04 \\
    \midrule
    ERM & 0.44 & 0.30 & 0.96 & 0.04 & 0.70 \\
    AutoAugment & 0.46 & 0.31 & 0.96 & 0.04 & 0.69 \\
    AugMix & 0.48 & 0.33 & 0.96 & 0.04 & 0.67 \\
    RandAugment & 0.51 & 0.37 & 0.94 & 0.06 & 0.63 \\
    \midrule
    Blurring & {\bf 0.74} & {\bf 0.98} & 0.49 & {\color{red} 0.51} & 0.02 \\
    Blurring + WP & \underline{0.58} & 0.47 & 0.87 & 0.13 & 0.53 \\
  \bottomrule
  \end{tabular}}
  \end{table}

\subsubsection{On the role of the input data: towards some practical recommendations regarding the training data}

Our results show that lowering the reliance on high-frequency content in the image improves generalization. This content is located on the 0.1-0.2 m scale and only appears on Google images. In \autoref{tab:flipped-experiments}, we flip our experiment to study how a model trained on IGN images generalizes to Google images. Results show that the model trained on IGN generalizes better to the downscaled Google images than the opposite. This result further supports the idea that higher GSD is not necessarily better for good robustness to acquisition conditions.

  \begin{table}[h]
\small
  \centering
  \vspace{0mm}\caption{\textbf{F1 Score} and true positives, true negatives, false positives, and false negatives rates. Evaluation computed on the Google dataset. ERM was trained on Google and Oracle on IGN images.}\label{tab:flipped-experiments}
  \resizebox{\textwidth}{!}{\begin{tabular}{r c c c c c}
  \toprule
   
   \multicolumn{1}{c}{Model}& F1 Score ($\uparrow$) & True positive rate ($\uparrow$) & True negative rate ($\uparrow$) & False positive rate ($\downarrow$) & False negative rate ($\downarrow$) \\
   \midrule
    ERM \citep{vapnik_nature_1999} &0.98&0,98 &	0.98&0,02&	0,02\\
    Oracle (ERM trained on IGN) & 0.91& 0,94 &	0,89&	0,11&	0,06\\
  \bottomrule
  \end{tabular}}
  \end{table}

%Paragraph on the noise and image quality for Google and IGN

%% file: content/conclusion.tex
\subsection{Conclusion}

This work is a comprehensive evaluation of the effects of distribution shifts on the classification accuracy of deep learning models trained to detect rooftop PV panels on overhead imagery. We first set up an experiment to disentangle the effects of the three primary shifts incurring in remote sensing \citep{tuia_domain_2016, murray_zoom_2019}, namely geographical variability, varying acquisition conditions, and varying ground sampling distance. We show that the varying acquisition conditions contribute significantly to the observed performance drop. To explain why this drop occurs, we leverage space-scale analysis to disentangle the different scales from the input images. We combine two types of explainable AI methods grounded in the wavelet decomposition of the input images to show that the CNN relies on noisy features (at the finest scales) and features that are not very well transferable across domains (at the coarsest scales). We then introduce a data augmentation technique to improve the model's robustness to distortions of the coarse-scale features and remove noise from the fine-scale features. We compare this method against various popular data augmentation techniques and show that our approach outperforms these baselines. We then discuss the practical takeaways of this work for the training or the choice of the training data for the initial training of the deep learning model. 

\paragraph{Broader impact} Currently, transmission system operators (TSOs) lack quality data regarding rooftop PV installations \citep{kasmi_towards_2022}. The lack of information leads to imprecise estimations and forecasts of the overall PV power generation, which in a context of sustained growth of the PV installed capacity could increase the uncertainty and threaten the grid's stability \citep{pierro_impact_2022}. On the other hand, current methods for mapping rooftop PV installations lack reliability, owing to their poor generalization abilities beyond their training dataset \citep{de_jong_monitoring_2020}. This work addresses this gap and thus demonstrates that remote sensing of PV installations is a reliable way for TSOs to improve their knowledge regarding small-scale PV installations and provides guidance on how to carry out the initial training of the model and how to deal with registry updates on newly available images.

\subsection{Limitations and future works}

\paragraph{Further discussion of the geographical variability} Our training data was limited to a narrow area around France. Therefore, we suspect the effect of the geographical variability to be underestimated. For instance, \cite{freitas_artificial_2023} showed that fine-tuning a model with data that is {\it not far} from the target area (e.g., France when the goal is to map PV systems in Portugal) enables accuracy gains compared to directly transferring a model trained over the United States. It could be interesting to study how the performance varies with the distance between the training data and the target mapping area once all other factors (acquisition conditions, ground sampling distance) are accounted for.

\paragraph{Extensions to other models} Over the last couple of years, foundation models \citep{bommasani_opportunities_2022} have been redefining the standards in deep learning. These very large models, trained on large data corpora, have shown remarkable performance for many challenging tasks, especially for text \citep{brown_language_2020} and image \citep{rombach_high-resolution_2022} generation. These models are used for more conventional and specialized tasks such as image segmentation \citep{kirillov_segment_2023} and achieve superior performance to conventional approaches while only requiring a few samples to learn their new task. Extending this benchmark and evaluating the performance of foundation models fine-tuned for segmenting PV panels such as \cite{yang_weakly-semi_2024} under distribution shifts could be interesting. 

%% file: content/appendix.tex
\section{Limitations of the GradCAM and related feature attribution methods for our use case}\label{sec:limitations}

\begin{figure}[h]
    \centering
    \includegraphics[width = \textwidth]{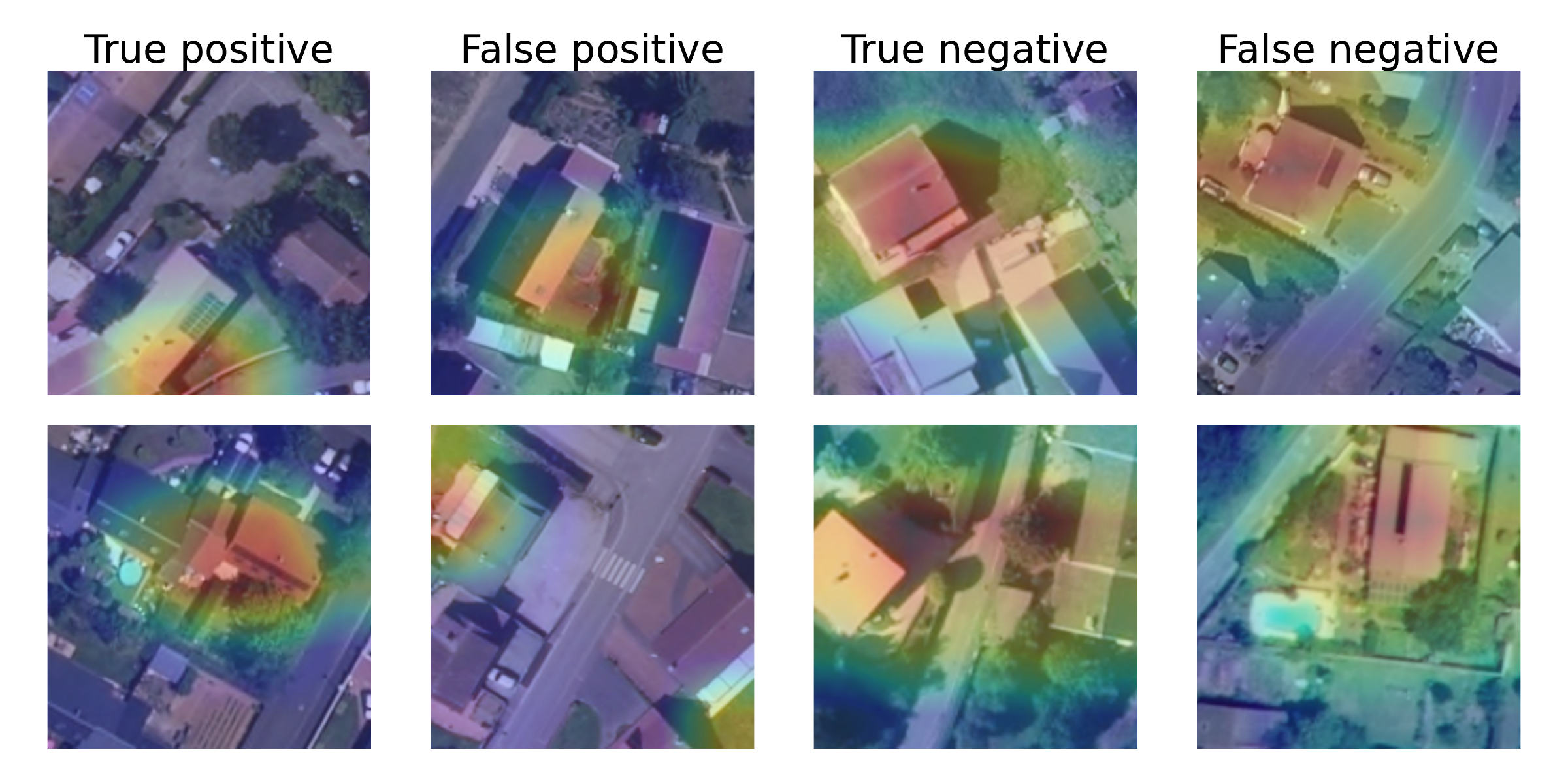}
    \caption{Model explanations using the GradCAM \citep{selvaraju_grad-cam_2020} for some true positives, false positives, true negatives and false negatives. The redder, the higher the contribution of an image region to the predicted class (1 for true and false positives, 0 for true and false negatives)}
    \label{fig:cam-explanations}
\end{figure}

\autoref{fig:cam-explanations} presents the explanations obtained using the GradCAM \citep{selvaraju_grad-cam_2020}. We can see two different prediction patterns depending on whether the model predicts a positive (true or false) or a negative (true or false). In the case of a true positive prediction, the model will focus on a specific, narrow region of the image, which indeed corresponds to a PV panel. However, for false positives, the model also focuses on a narrow image region. Inspecting the samples of \autoref{fig:cam-explanations} reveals that this region of the image depicts items that {\it resemble} PV panels. On the image on the first row (second column) of \autoref{fig:cam-explanations}, we can see that the model confuses a shadehouse that shares the same color and overall shape of a PV panel with an actual panel. In the image on the second row, the verandas with groves fool the model.

On the other hand, when the model does not see a PV panel, it does not focus on a specific image region. This remains true for the false negatives, where we can see that the model does not see the panels on any of the images. 

However, we can also see that as the GradCAM only assesses where the model is looking, it is challenging to understand {\ why} it focused on a given area that resembles a PV panel on false positives and why it did not identify the PV panel on the false negatives. \cite{achtibat_where_2022} underlines the necessity for reliable model evaluation to assess where models are looking at and {\it what} they are looking at on input images. The choice of the WCAM as an attribution method and, more broadly, the space-scale decomposition is an attempt to address this question by assessing the scales the models consider when making their predictions.

\section{Computation of the WCAM (from \cite{kasmi_assessment_2023})}\label{sec:wcam-computation}

\begin{figure}[H]
\small
    \centering
    \includegraphics[width = \textwidth]{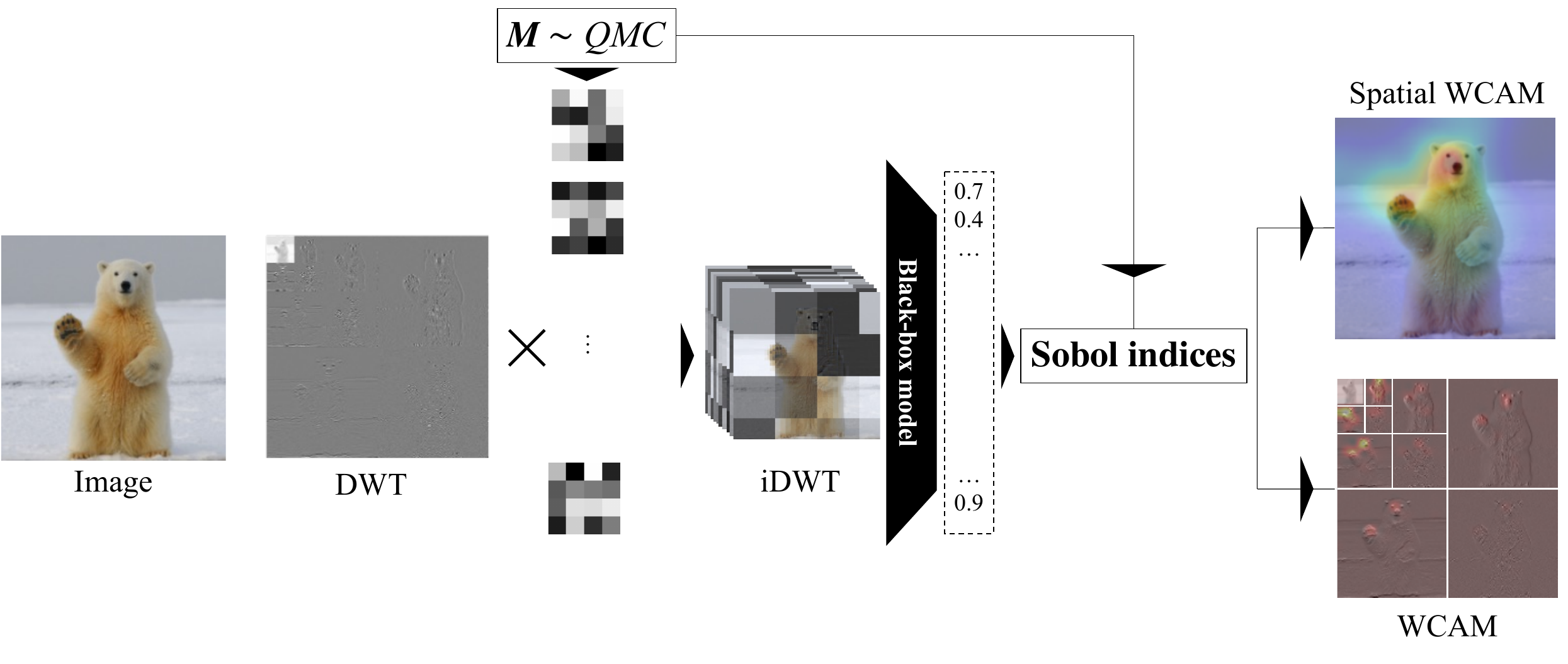}
    \caption{Flowchart of the wavelet scale attribution method (WCAM). Source: \cite{kasmi_assessment_2023}}
    \label{fig:wcam-principle}
\end{figure}

\autoref{fig:wcam-principle} depicts the principle of the WCAM. The importance of the regions of the wavelet transform of the input image is estimated by {\bf (1)} generating masks from a Quasi-Monte Carlo sequence, {\bf (2)} evaluating the model on perturbed images. We obtain these images by computing the discrete wavelet transform (DWT) of the original image, applying the masks on the DWT to obtain perturbed DWT,\footnote{On an RGB image, we apply the DWT channel-wise and apply the same perturbation to each channel.} and inverting the perturbed DWT to generate perturbed images. We generate $N(K+2)$ perturbed images for a single image. {\bf (3)} We estimate the total Sobol indices of the perturbed regions of the wavelet transform using the masks and the model's outputs using Jansen's estimator \citep{jansen_analysis_1999}. \cite{fel_look_2021} introduced this approach to estimate the importance of image regions in the pixel space. We generalize it to the wavelet domain.

\section{Quantitative relationship between the WCAM's scale embeddings and the model's response to distribution shifts}\label{sec:scale-embeddings}

\paragraph{Definition} 

A {\it scale embedding} is a vector $z = (z_1,\dots, z_L)\in\mathbb{R}^L$ where each component $z_s$ encodes the importance of the $l^{\textrm{th}}$ scale component in the prediction.     

Scale embeddings compute the importance of each scale and each direction and summarize it into a vector $z\in\mathbb{R}^L$ where $L$ indicates the number of levels. In our case, we have ten levels (1 corresponding to the approximation coefficients and $L=$ 3 $\times$ 3 corresponding to the three scales of details coefficients and their three respective orientations. Scale embeddings summarize the importance of each scale, irrespective of the spatial localization of importance

\paragraph{Results} We computed the distance (measured by the Euclidean distance) between two images' scale embeddings and computed the correlation between this distance and the predicted probability shift. As a baseline, we also computed the distance between the two WCAMs.

We obtained correlation coefficients of 0.18 ($p=0.19$) for the scale embedding and 0.17 ($p=0.19$) for the raw WCAM. Although weaker than the correlation between the distortion and the predicted probability shift, this result highlights that the WCAM consistently captures the change in behavior of the model resulting from the shift in acquisition conditions.
\section{Data augmentation strategies}

\subsection{Description of the data augmentations}\label{sec:augmentation-plot}

\paragraph{AugMix \citep{hendrycks_augmix_2020}} The data augmentation strategy "Augment-and-Mix" (AugMix) consists of producing a high diversity of augmented images from an input sample. A set of operations (perturbations) to be applied to the images are sampled, along with sampling weights. The image resulting $x_{aug}$ is obtained through the composition $x_{aug} = \omega_1 op_1 \circ \dots \omega_n op_n (x) $ where $x$ is the original image. Then, the augmented image is interpolated with the original image with a weight $m$ that is also randomly sampled. We have $x_{augmix} = mx + (1-m)x_{aug}$. 

\paragraph{AutoAugment \citep{cubuk_autoaugment_2019}} This strategy aims at finding the best data augmentation for a given dataset. The authors determined the best augmentations strategy $S$ as the outcome of a reinforcement learning problem: a controller predicts an augmentation policy from a search space. Then, the authors train a model, and the controller updates its sampling strategy $S$ based on the train loss. The goal is for the controller to generate better policies over time. The authors derive optimal augmentation strategies for various datasets, including ImageNet \citep{russakovsky_imagenet_2015}, and show that the optimal policy for ImageNet generalizes well to other datasets.

\paragraph{RandAugment \citep{cubuk_randaugment_2020}} This strategy's primary goal is to remove the need for a computationally expansive policy search before model training. Instead of searching for transformations, random probabilities are assigned to the transformations. Then, each resulting policy (a weighted sequence of $K$ transformations) is graded depending on its strength. The number of transformations and the strength are passed as input when calling the transformation. 

\paragraph{Blurring} We apply a nonrandom Gaussian blur to the image. The value is set by comparing visually Google and IGN images and trying to remove details from Google images that are not visible on IGN images. After a manual inspection, we set the blur level to discard the details at 0.1-0.2 m scale from the image. It corresponds to a blurring value $\sigma = 2.$ in the {\tt ImageFilter.GaussianBlur} method of the Python Imaging Library (PIL). 

\paragraph{Blurring + Wavelet perturbation (WP)} We first blur the image. Then, for each color channel, we compute the dyadic wavelet transform of the image and randomly perturb the coefficients (we randomly set some coefficients to 0) at targeted scales. The set of coefficients set to 0 is determined with uniform sampling. This results in a random perturbation that removes information for some precise scales and locations. We then reconstruct the image from its perturbed wavelet coefficients. For each call, 20\% of the coefficients are canceled. This value balances between the loss of information and the input perturbation. We perturb each color channel independently. The wavelet perturbation aims to disrupt information at specific scales, as it can happen with varying acquisition conditions.

%% faire un algorithme ici

\subsection{Plots}\label{sec:plot-augmentation}

\begin{figure}[h]
    \centering
    \includegraphics[width = \textwidth]{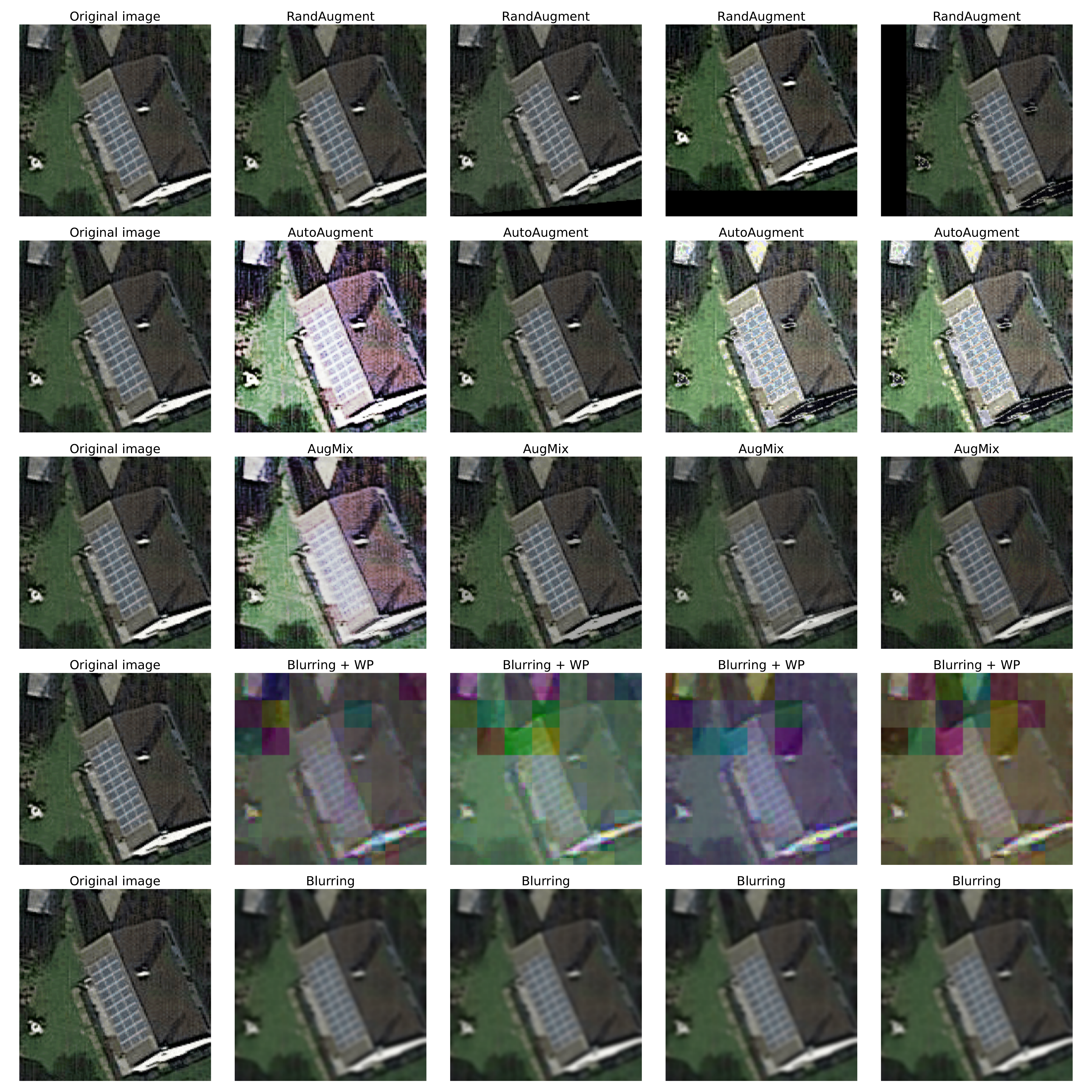}
    \caption{Visualization of the different data augmentation techniques implemented in this work}
    \label{fig:augmentations}
\end{figure}

\autoref{fig:augmentations} plots examples of the different data augmentations implemented in this work. Along with these augmentations, we apply random rotations, symmetries, and normalization to the input during training. At test time, we only normalize the input images.

\newpage
\section{Accuracy results for variants of the Scattering transform}\label{sec:appendix-stm}

\autoref{tab:results-scattering-full} presents the accuracy of the Scattering transform for two depth variants (labeled $m=1$ and $m=2$). We can see that the performance of the Scattering transform remains relatively poor regardless of the depth of the scattering coefficients. Contrary to the claims of \cite{bruna_invariant_2013}, including second-order coefficients does not seem enough to discriminate between images, as the number of false positives remains high. This could be caused by the fact that our task, namely the detection of small objects on overhead imagery, is more challenging than digit classifications. 

  \begin{table}[h]
\small
  \centering
  \setlength{\tabcolsep}{2pt}
  \vspace{0mm}\caption{{\bf F1 Score} and decomposition in true positives, true negatives, false positives and false negative rate of the classification accuracy of the Scattering Transform model trained on Google images and deployed on IGN images. }\label{tab:results-scattering-full}
  \resizebox{\textwidth}{!}{\begin{tabular}{c r c c c c c}
  \toprule
    Depth&\multicolumn{1}{c}{Dataset}&  F1 Score ($\uparrow$) & True positive rate ($\uparrow$) & True negative rate ($\uparrow$) & False positive rate ($\downarrow$) & False negative rate ($\downarrow$) \\
    \midrule
        \multirow{2}{*}{$m=1$}&{\it Google baseline}&0.57&0,84	&0,09	&0,57	&0,57\\
                        &IGN&0.57&0,71	&0,40	&0,52	&0,36\\
                        \cmidrule{2-7}
    \multirow{2}{*}{$m=2$}&{\it Google baseline}&0.57&0,89	&0,10	&0,56	&0,48\\
                        &IGN&0.59&0,54	&0,31	&0,62	&0,54\\
                        \cmidrule{2-7}
    %\multirow{2}{*}{$m=3$}&{\it Google baseline}&0.45&0,91	&0,08	&0,55	&0,45\\
    %                    &IGN&0.60&0,68&	0,44&	0,51&	0,36\\
    %\midrule
    \multirow{2}{*}{{\it ERM}}& {\it Google baseline} &  0.98  & 0.99 & 0.98 & 0.02& 0.01 \\ 
    &{\it IGN} & 0.46 & 0.32 & 0.95 & 0.03& 0.68\\
    \cmidrule{2-7}
    \multirow{2}{*}{{\it Random classifier}}&{\it Google baseline}&0.47 & 0,5&	0,50&	0,55&	0,45\\
    &{\it IGN}&0.47 & 0,50&	0,50&	0,56&	0,44\\
  \bottomrule
  \end{tabular}}
  \end{table}
\newpage
\section{Additional figures}\label{sec:additional-figues}

\begin{figure}[H]
    \centering
    \includegraphics[width=.6\textwidth]{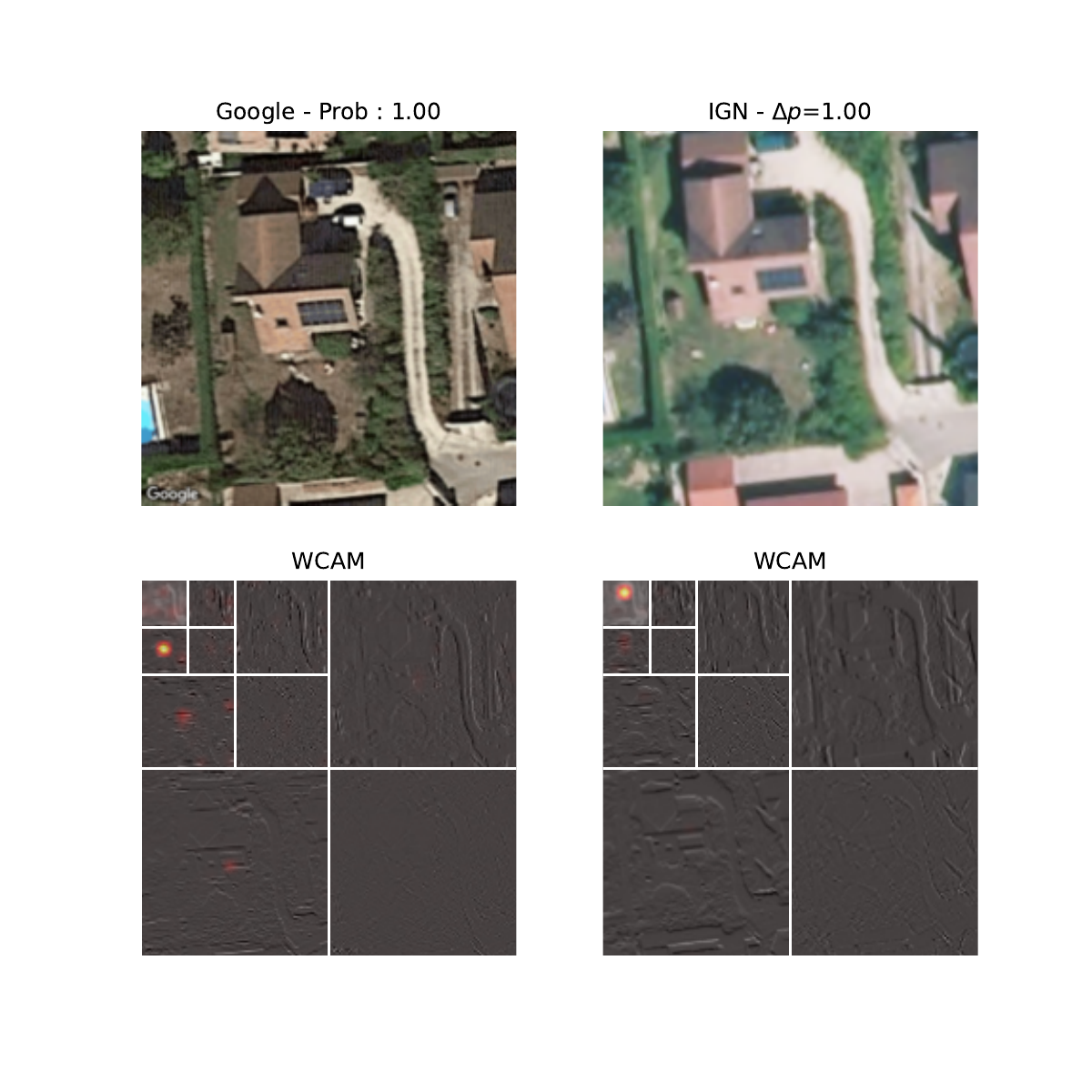}
    \caption{Analysis with the WCAM of the CNNs prediction on an image no longer recognized as a PV panel}
    \label{fig:wcam-example-2}
\end{figure}

\begin{figure}[H]
    \centering
    \includegraphics[width=.6\textwidth]{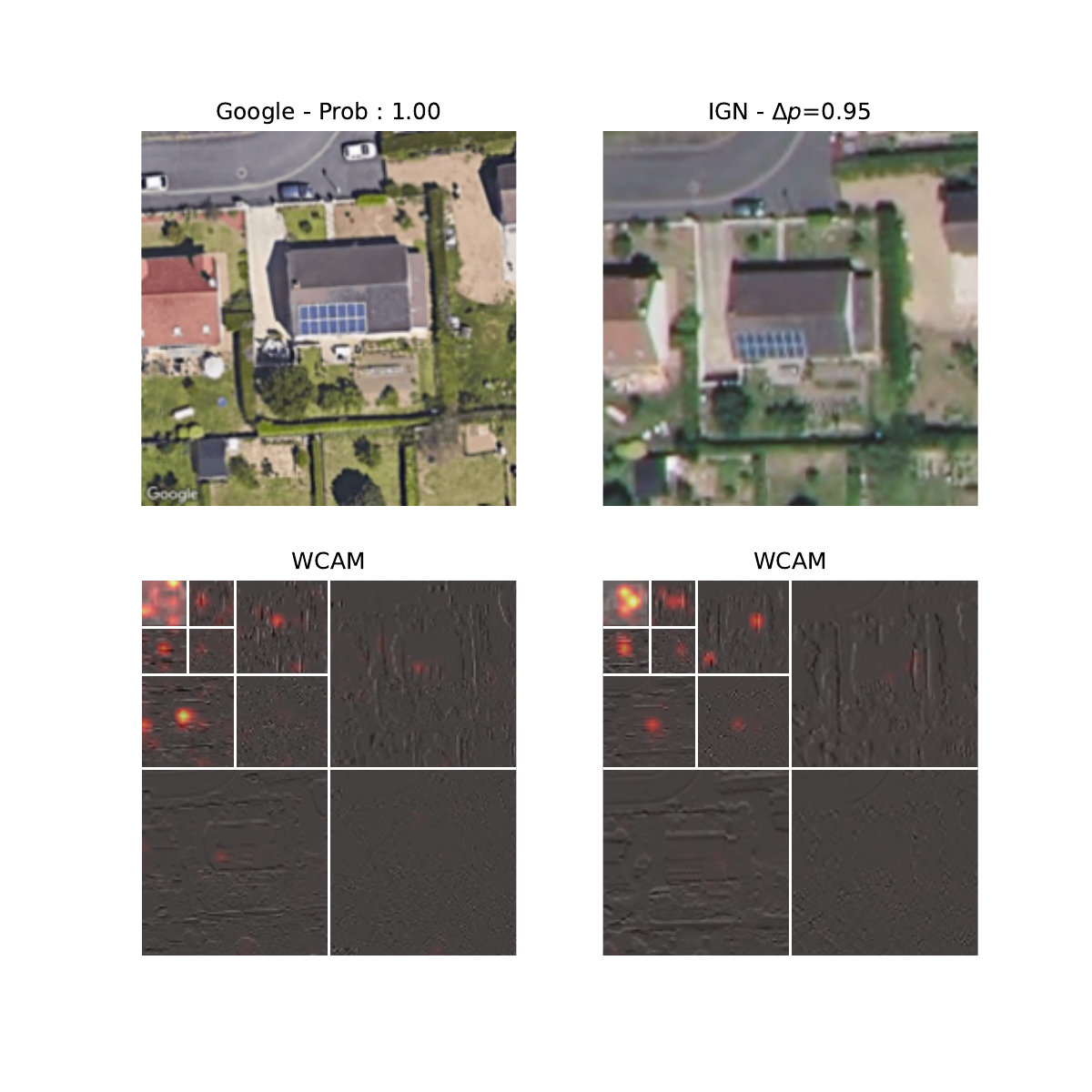}
    \caption{Analysis with the WCAM of the CNNs prediction on an image no longer recognized as a PV panel}
    \label{fig:wcam-example-3}
\end{figure}

\begin{figure}[H]
    \centering
    \includegraphics[width=.6\textwidth]{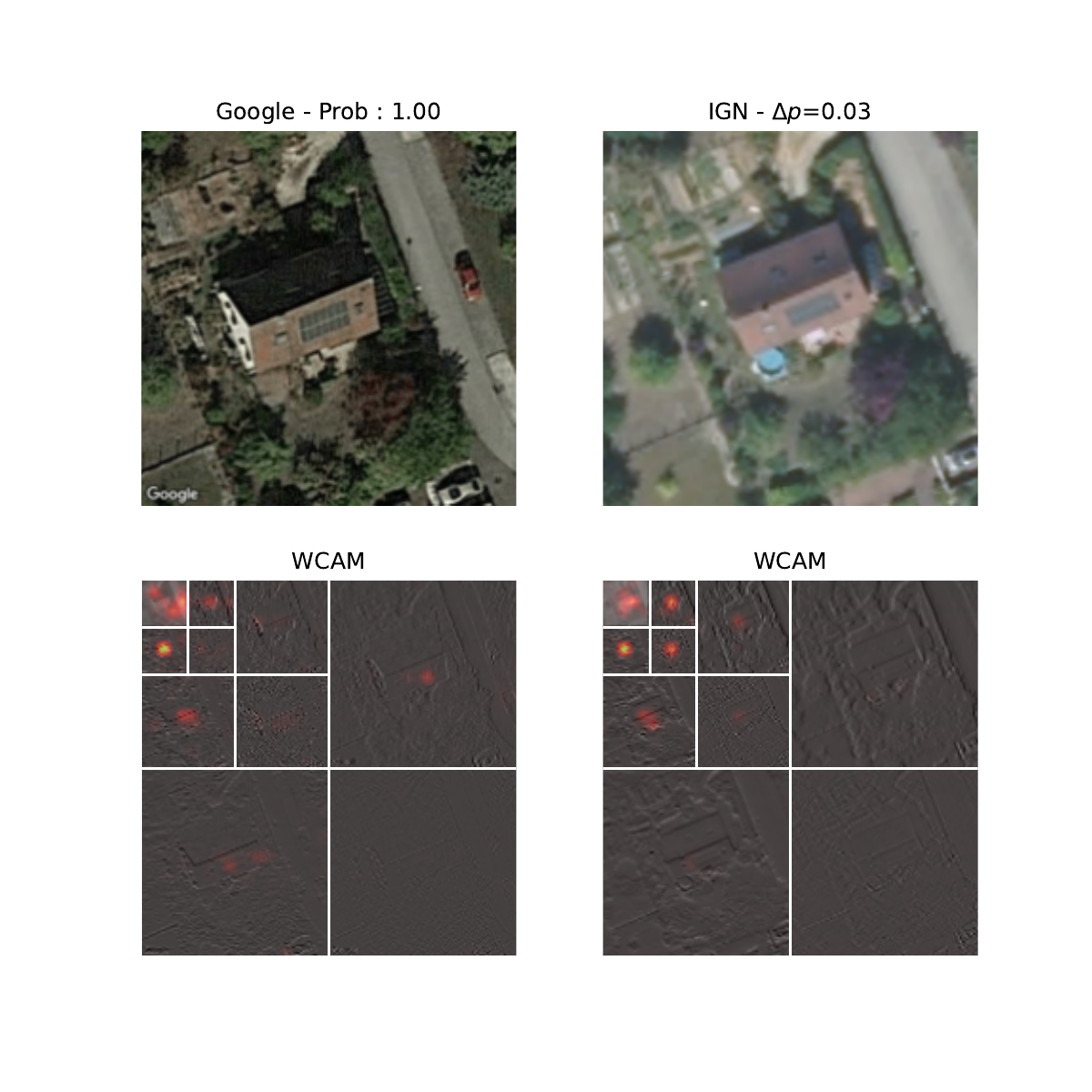}
    \caption{Analysis with the WCAM of the CNNs prediction on an image that is insensitive to varying acquisition conditions}
    \label{fig:wcam-example-4}
\end{figure}

\autoref{fig:wcam-example-2} to \autoref{fig:wcam-example-4} present additional examples of qualitative assessment of the effects of distribution shifts on the model's prediction. On \autoref{fig:wcam-example-2}, we can see that the model initially primarily relied on the gridded pattern, particularly visible at the 4-8 pixel scale. The acquisition conditions discarded this factor, thus explaining why the model could no longer recognize the PV panel. A similar phenomenon occurs on \autoref{fig:wcam-example-3}. \autoref{fig:wcam-example-4} presents an example of a prediction not affected by the acquisition conditions. We can see that the important scales (especially at the 4-8 pixel scale) remain the same.

%% file: references.bib
@book{pfungst_clever_1911,
	title = {Clever {Hans}:(the horse of {Mr}. {Von} {Osten}.) a contribution to experimental animal and human psychology},
	publisher = {Holt, Rinehart and Winston},
	author = {Pfungst, Oskar},
	year = {1911},
}

@article{wang_image_2004,
	title = {Image quality assessment: from error visibility to structural similarity},
	volume = {13},
	number = {4},
	journal = {IEEE transactions on image processing},
	author = {Wang, Zhou and Bovik, Alan C and Sheikh, Hamid R and Simoncelli, Eero P},
	year = {2004},
	note = {Publisher: IEEE},
	pages = {600--612},
}

@article{lone_noise_2018,
	title = {Noise models in digital image processing},
	volume = {10},
	number = {2},
	journal = {Global Sci-Tech},
	author = {Lone, Aamir Hamid and Siddiqui, Arsheen Neda},
	year = {2018},
	note = {Publisher: Al-falah School of Engineering and Technology},
	pages = {63--66},
}

@misc{brown_language_2020,
	title = {Language {Models} are {Few}-{Shot} {Learners}},
	url = {http://arxiv.org/abs/2005.14165},
	doi = {10.48550/arXiv.2005.14165},
	urldate = {2024-06-18},
	publisher = {arXiv},
	author = {Brown, Tom B. and Mann, Benjamin and Ryder, Nick and Subbiah, Melanie and Kaplan, Jared and Dhariwal, Prafulla and Neelakantan, Arvind and Shyam, Pranav and Sastry, Girish and Askell, Amanda and Agarwal, Sandhini and Herbert-Voss, Ariel and Krueger, Gretchen and Henighan, Tom and Child, Rewon and Ramesh, Aditya and Ziegler, Daniel M. and Wu, Jeffrey and Winter, Clemens and Hesse, Christopher and Chen, Mark and Sigler, Eric and Litwin, Mateusz and Gray, Scott and Chess, Benjamin and Clark, Jack and Berner, Christopher and McCandlish, Sam and Radford, Alec and Sutskever, Ilya and Amodei, Dario},
	month = jul,
	year = {2020},
	note = {arXiv:2005.14165 [cs]},
}

@misc{kirillov_segment_2023,
	title = {Segment {Anything}},
	url = {http://arxiv.org/abs/2304.02643},
	doi = {10.48550/arXiv.2304.02643},
	urldate = {2024-06-18},
	publisher = {arXiv},
	author = {Kirillov, Alexander and Mintun, Eric and Ravi, Nikhila and Mao, Hanzi and Rolland, Chloe and Gustafson, Laura and Xiao, Tete and Whitehead, Spencer and Berg, Alexander C. and Lo, Wan-Yen and Dollár, Piotr and Girshick, Ross},
	month = apr,
	year = {2023},
	note = {arXiv:2304.02643 [cs]},
}

@inproceedings{rombach_high-resolution_2022,
	title = {High-{Resolution} {Image} {Synthesis} {With} {Latent} {Diffusion} {Models}},
	booktitle = {Proceedings of the {IEEE}/{CVF} {Conference} on {Computer} {Vision} and {Pattern} {Recognition} ({CVPR})},
	author = {Rombach, Robin and Blattmann, Andreas and Lorenz, Dominik and Esser, Patrick and Ommer, Björn},
	month = jun,
	year = {2022},
	pages = {10684--10695},
}

@article{andreux_kymatio_2020,
	title = {Kymatio: {Scattering} {Transforms} in {Python}},
	volume = {21},
	url = {http://jmlr.org/papers/v21/19-047.html},
	number = {60},
	journal = {Journal of Machine Learning Research},
	author = {Andreux, Mathieu and Angles, Tomás and Exarchakis, Georgios and Leonarduzzi, Roberto and Rochette, Gaspar and Thiry, Louis and Zarka, John and Mallat, Stéphane and Andén, Joakim and Belilovsky, Eugene and Bruna, Joan and Lostanlen, Vincent and Chaudhary, Muawiz and Hirn, Matthew J. and Oyallon, Edouard and Zhang, Sixin and Cella, Carmine and Eickenberg, Michael},
	year = {2020},
	pages = {1--6},
}

@misc{sudjianto_designing_2021,
	title = {Designing {Inherently} {Interpretable} {Machine} {Learning} {Models}},
	url = {http://arxiv.org/abs/2111.01743},
	language = {en},
	urldate = {2024-06-17},
	publisher = {arXiv},
	author = {Sudjianto, Agus and Zhang, Aijun},
	month = nov,
	year = {2021},
	note = {arXiv:2111.01743 [cs, stat]},
}

@phdthesis{parekh_cadre_2023,
	type = {Theses},
	title = {Un cadre flexible pour l'apprentissage automatique interprétable : application à la classification d'images et d'audio},
	url = {https://theses.hal.science/tel-04214919},
	school = {Institut Polytechnique de Paris},
	author = {Parekh, Jayneel},
	month = jul,
	year = {2023},
	note = {Issue: 2023IPPAT032},
}

@article{hu_what_2022,
	title = {What you get is not always what you see—pitfalls in solar array assessment using overhead imagery},
	volume = {327},
	issn = {0306-2619},
	url = {https://www.sciencedirect.com/science/article/pii/S0306261922014003},
	doi = {10.1016/j.apenergy.2022.120143},
	urldate = {2024-05-16},
	journal = {Applied Energy},
	author = {Hu, Wei and Bradbury, Kyle and Malof, Jordan M. and Li, Boning and Huang, Bohao and Streltsov, Artem and Sydny Fujita, K. and Hoen, Ben},
	month = dec,
	year = {2022},
	pages = {120143},
}

@inproceedings{fel_look_2021,
	title = {Look at the {Variance}! {Efficient} {Black}-box {Explanations} with {Sobol}-based {Sensitivity} {Analysis}},
	volume = {34},
	url = {https://proceedings.neurips.cc/paper_files/paper/2021/file/da94cbeff56cfda50785df477941308b-Paper.pdf},
	booktitle = {Advances in {Neural} {Information} {Processing} {Systems}},
	publisher = {Curran Associates, Inc.},
	author = {Fel, Thomas and Cadene, Remi and Chalvidal, Mathieu and Cord, Matthieu and Vigouroux, David and Serre, Thomas},
	editor = {Ranzato, M. and Beygelzimer, A. and Dauphin, Y. and Liang, P. S. and Vaughan, J. Wortman},
	year = {2021},
	pages = {26005--26014},
}

@article{selvaraju_grad-cam_2020,
	title = {Grad-{CAM}: {Visual} {Explanations} from {Deep} {Networks} via {Gradient}-{Based} {Localization}},
	volume = {128},
	issn = {1573-1405},
	shorttitle = {Grad-{CAM}},
	url = {https://doi.org/10.1007/s11263-019-01228-7},
	doi = {10.1007/s11263-019-01228-7},
	language = {en},
	number = {2},
	urldate = {2024-05-14},
	journal = {International Journal of Computer Vision},
	author = {Selvaraju, Ramprasaath R. and Cogswell, Michael and Das, Abhishek and Vedantam, Ramakrishna and Parikh, Devi and Batra, Dhruv},
	month = feb,
	year = {2020},
	pages = {336--359},
}

@inproceedings{dardouillet_explainability_2023,
	address = {Cham},
	title = {Explainability of {Image} {Semantic} {Segmentation} {Through} {SHAP} {Values}},
	isbn = {978-3-031-37731-0},
	doi = {10.1007/978-3-031-37731-0_19},
	language = {en},
	booktitle = {Pattern {Recognition}, {Computer} {Vision}, and {Image} {Processing}. {ICPR} 2022 {International} {Workshops} and {Challenges}},
	publisher = {Springer Nature Switzerland},
	author = {Dardouillet, Pierre and Benoit, Alexandre and Amri, Emna and Bolon, Philippe and Dubucq, Dominique and Credoz, Anthony},
	editor = {Rousseau, Jean-Jacques and Kapralos, Bill},
	year = {2023},
	pages = {188--202},
}

@article{zhou_domain_2023,
	title = {Domain {Generalization}: {A} {Survey}},
	volume = {45},
	issn = {1939-3539},
	shorttitle = {Domain {Generalization}},
	url = {https://ieeexplore.ieee.org/abstract/document/9847099},
	doi = {10.1109/TPAMI.2022.3195549},
	number = {4},
	urldate = {2024-05-14},
	journal = {IEEE Transactions on Pattern Analysis and Machine Intelligence},
	author = {Zhou, Kaiyang and Liu, Ziwei and Qiao, Yu and Xiang, Tao and Loy, Chen Change},
	month = apr,
	year = {2023},
	note = {Conference Name: IEEE Transactions on Pattern Analysis and Machine Intelligence},
	pages = {4396--4415},
}

@article{guan_domain_2022,
	title = {Domain {Adaptation} for {Medical} {Image} {Analysis}: {A} {Survey}},
	volume = {69},
	issn = {1558-2531},
	shorttitle = {Domain {Adaptation} for {Medical} {Image} {Analysis}},
	url = {https://ieeexplore.ieee.org/document/9557808},
	doi = {10.1109/TBME.2021.3117407},
	number = {3},
	urldate = {2024-05-13},
	journal = {IEEE Transactions on Biomedical Engineering},
	author = {Guan, Hao and Liu, Mingxia},
	month = mar,
	year = {2022},
	note = {Conference Name: IEEE Transactions on Biomedical Engineering},
	pages = {1173--1185},
}

@inproceedings{casanova_instance-conditioned_2021,
	title = {Instance-{Conditioned} {GAN}},
	volume = {34},
	url = {https://proceedings.neurips.cc/paper_files/paper/2021/file/e7ac288b0f2d41445904d071ba37aaff-Paper.pdf},
	booktitle = {Advances in {Neural} {Information} {Processing} {Systems}},
	publisher = {Curran Associates, Inc.},
	author = {Casanova, Arantxa and Careil, Marlene and Verbeek, Jakob and Drozdzal, Michal and Romero Soriano, Adriana},
	editor = {Ranzato, M. and Beygelzimer, A. and Dauphin, Y. and Liang, P. S. and Vaughan, J. Wortman},
	year = {2021},
	pages = {27517--27529},
}

@misc{csurka_unsupervised_2021,
	title = {Unsupervised {Domain} {Adaptation} for {Semantic} {Image} {Segmentation}: a {Comprehensive} {Survey}},
	shorttitle = {Unsupervised {Domain} {Adaptation} for {Semantic} {Image} {Segmentation}},
	url = {http://arxiv.org/abs/2112.03241},
	doi = {10.48550/arXiv.2112.03241},
	urldate = {2024-05-13},
	publisher = {arXiv},
	author = {Csurka, Gabriela and Volpi, Riccardo and Chidlovskii, Boris},
	month = dec,
	year = {2021},
	note = {arXiv:2112.03241 [cs]},
}

@incollection{csurka_comprehensive_2017,
	address = {Cham},
	title = {A {Comprehensive} {Survey} on {Domain} {Adaptation} for {Visual} {Applications}},
	isbn = {978-3-319-58347-1},
	url = {https://doi.org/10.1007/978-3-319-58347-1_1},
	language = {en},
	urldate = {2024-05-13},
	booktitle = {Domain {Adaptation} in {Computer} {Vision} {Applications}},
	publisher = {Springer International Publishing},
	author = {Csurka, Gabriela},
	editor = {Csurka, Gabriela},
	year = {2017},
	doi = {10.1007/978-3-319-58347-1_1},
	pages = {1--35},
}

@inproceedings{hendrycks_benchmarking_2019,
	title = {Benchmarking {Neural} {Network} {Robustness} to {Common} {Corruptions} and {Perturbations}},
	url = {https://openreview.net/forum?id=HJz6tiCqYm},
	booktitle = {International {Conference} on {Learning} {Representations}},
	author = {Hendrycks, Dan and Dietterich, Thomas},
	year = {2019},
}

@inproceedings{simonyan_very_2015,
	title = {Very {Deep} {Convolutional} {Networks} for {Large}-{Scale} {Image} {Recognition}},
	url = {http://arxiv.org/abs/1409.1556},
	booktitle = {3rd {International} {Conference} on {Learning} {Representations}, {ICLR} 2015, {San} {Diego}, {CA}, {USA}, {May} 7-9, 2015, {Conference} {Track} {Proceedings}},
	author = {Simonyan, Karen and Zisserman, Andrew},
	editor = {Bengio, Yoshua and LeCun, Yann},
	year = {2015},
}

@inproceedings{recht_imagenet_2019,
	title = {Do {ImageNet} {Classifiers} {Generalize} to {ImageNet}?},
	url = {https://proceedings.mlr.press/v97/recht19a.html},
	language = {en},
	urldate = {2024-05-13},
	booktitle = {Proceedings of the 36th {International} {Conference} on {Machine} {Learning}},
	publisher = {PMLR},
	author = {Recht, Benjamin and Roelofs, Rebecca and Schmidt, Ludwig and Shankar, Vaishaal},
	month = may,
	year = {2019},
	note = {ISSN: 2640-3498},
	pages = {5389--5400},
}

@inproceedings{geirhos_imagenet-trained_2019,
	title = {{ImageNet}-trained {CNNs} are biased towards texture; increasing shape bias improves accuracy and robustness.},
	url = {https://openreview.net/forum?id=Bygh9j09KX},
	booktitle = {International {Conference} on {Learning} {Representations}},
	author = {Geirhos, Robert and Rubisch, Patricia and Michaelis, Claudio and Bethge, Matthias and Wichmann, Felix A. and Brendel, Wieland},
	year = {2019},
}

@article{russakovsky_imagenet_2015,
	title = {{ImageNet} {Large} {Scale} {Visual} {Recognition} {Challenge}},
	volume = {115},
	issn = {1573-1405},
	url = {https://doi.org/10.1007/s11263-015-0816-y},
	doi = {10.1007/s11263-015-0816-y},
	language = {en},
	number = {3},
	urldate = {2024-05-13},
	journal = {International Journal of Computer Vision},
	author = {Russakovsky, Olga and Deng, Jia and Su, Hao and Krause, Jonathan and Satheesh, Sanjeev and Ma, Sean and Huang, Zhiheng and Karpathy, Andrej and Khosla, Aditya and Bernstein, Michael and Berg, Alexander C. and Fei-Fei, Li},
	month = dec,
	year = {2015},
	pages = {211--252},
}

@inproceedings{petsiuk_rise_2018,
	title = {{RISE}: {Randomized} {Input} {Sampling} for {Explanation} of {Black}-box {Models}},
	shorttitle = {{RISE}},
	url = {http://arxiv.org/abs/1806.07421},
	doi = {10.48550/arXiv.1806.07421},
	urldate = {2023-03-30},
	author = {Petsiuk, Vitali and Das, Abir and Saenko, Kate},
	month = sep,
	year = {2018},
	note = {arXiv:1806.07421 [cs]},
}

@inproceedings{selvaraju_grad-cam_2017,
	title = {Grad-{CAM}: {Visual} {Explanations} from {Deep} {Networks} via {Gradient}-{Based} {Localization}},
	shorttitle = {Grad-{CAM}},
	url = {https://ieeexplore.ieee.org/document/8237336},
	doi = {10.1109/ICCV.2017.74},
	urldate = {2024-05-13},
	booktitle = {2017 {IEEE} {International} {Conference} on {Computer} {Vision} ({ICCV})},
	author = {Selvaraju, Ramprasaath R. and Cogswell, Michael and Das, Abhishek and Vedantam, Ramakrishna and Parikh, Devi and Batra, Dhruv},
	month = oct,
	year = {2017},
	note = {ISSN: 2380-7504},
	pages = {618--626},
}

@inproceedings{zhang_understanding_2017,
	title = {Understanding deep learning requires rethinking generalization},
	url = {https://openreview.net/forum?id=Sy8gdB9xx},
	booktitle = {5th {International} {Conference} on {Learning} {Representations}, {ICLR} 2017, {Toulon}, {France}, {April} 24-26, 2017, {Conference} {Track} {Proceedings}},
	publisher = {OpenReview.net},
	author = {Zhang, Chiyuan and Bengio, Samy and Hardt, Moritz and Recht, Benjamin and Vinyals, Oriol},
	year = {2017},
}

@article{tuia_domain_2016,
	title = {Domain {Adaptation} for the {Classification} of {Remote} {Sensing} {Data}: {An} {Overview} of {Recent} {Advances}},
	volume = {4},
	issn = {2168-6831},
	shorttitle = {Domain {Adaptation} for the {Classification} of {Remote} {Sensing} {Data}},
	url = {https://ieeexplore.ieee.org/document/7486184},
	doi = {10.1109/MGRS.2016.2548504},
	number = {2},
	urldate = {2024-05-11},
	journal = {IEEE Geoscience and Remote Sensing Magazine},
	author = {Tuia, Devis and Persello, Claudio and Bruzzone, Lorenzo},
	month = jun,
	year = {2016},
	note = {Conference Name: IEEE Geoscience and Remote Sensing Magazine},
	pages = {41--57},
}

@inproceedings{torralba_unbiased_2011,
	title = {Unbiased look at dataset bias},
	url = {https://ieeexplore.ieee.org/document/5995347},
	doi = {10.1109/CVPR.2011.5995347},
	urldate = {2024-05-11},
	booktitle = {{CVPR} 2011},
	author = {Torralba, Antonio and Efros, Alexei A.},
	month = jun,
	year = {2011},
	note = {ISSN: 1063-6919},
	pages = {1521--1528},
}

@misc{de_jong_monitoring_2020,
	title = {Monitoring {Spatial} {Sustainable} {Development}: semi-automated analysis of {Satellite} and {Aerial} {Images} for {Energy} {Transition} and {Sustainability} {Indicators}},
	doi = {https://doi.org/10.48550/arXiv.2009.05738},
	publisher = {arXiv},
	author = {De Jong, Tim and Bromuri, Stefano and Chang, Xi and Debusschere, Marc and Rosenski, Natalie and Schartner, Clara and Strauch, Katharina and Boehmer, Marion and Curier, Lyana},
	year = {2020},
}

@inproceedings{koh_wilds_2021,
	title = {{WILDS}: {A} {Benchmark} of in-the-{Wild} {Distribution} {Shifts}},
	shorttitle = {{WILDS}},
	url = {https://proceedings.mlr.press/v139/koh21a.html},
	language = {en},
	urldate = {2024-05-11},
	booktitle = {Proceedings of the 38th {International} {Conference} on {Machine} {Learning}},
	publisher = {PMLR},
	author = {Koh, Pang Wei and Sagawa, Shiori and Marklund, Henrik and Xie, Sang Michael and Zhang, Marvin and Balsubramani, Akshay and Hu, Weihua and Yasunaga, Michihiro and Phillips, Richard Lanas and Gao, Irena and Lee, Tony and David, Etienne and Stavness, Ian and Guo, Wei and Earnshaw, Berton and Haque, Imran and Beery, Sara M. and Leskovec, Jure and Kundaje, Anshul and Pierson, Emma and Levine, Sergey and Finn, Chelsea and Liang, Percy},
	month = jul,
	year = {2021},
	note = {ISSN: 2640-3498},
	pages = {5637--5664},
}

@article{bruna_invariant_2013,
	title = {Invariant {Scattering} {Convolution} {Networks}},
	volume = {35},
	issn = {1939-3539},
	url = {https://ieeexplore.ieee.org/document/6522407},
	doi = {10.1109/TPAMI.2012.230},
	number = {8},
	urldate = {2024-05-11},
	journal = {IEEE Transactions on Pattern Analysis and Machine Intelligence},
	author = {Bruna, Joan and Mallat, Stephane},
	month = aug,
	year = {2013},
	note = {Conference Name: IEEE Transactions on Pattern Analysis and Machine Intelligence},
	pages = {1872--1886},
}

@inproceedings{cubuk_autoaugment_2019,
	title = {{AutoAugment}: {Learning} {Augmentation} {Strategies} {From} {Data}},
	booktitle = {Proceedings of the {IEEE}/{CVF} {Conference} on {Computer} {Vision} and {Pattern} {Recognition} ({CVPR})},
	author = {Cubuk, Ekin D. and Zoph, Barret and Mane, Dandelion and Vasudevan, Vijay and Le, Quoc V.},
	month = jun,
	year = {2019},
}

@inproceedings{lundberg_unified_2017,
	title = {A {Unified} {Approach} to {Interpreting} {Model} {Predictions}},
	volume = {30},
	url = {https://proceedings.neurips.cc/paper_files/paper/2017/file/8a20a8621978632d76c43dfd28b67767-Paper.pdf},
	booktitle = {Advances in {Neural} {Information} {Processing} {Systems}},
	publisher = {Curran Associates, Inc.},
	author = {Lundberg, Scott M and Lee, Su-In},
	editor = {Guyon, I. and Luxburg, U. Von and Bengio, S. and Wallach, H. and Fergus, R. and Vishwanathan, S. and Garnett, R.},
	year = {2017},
}

@inproceedings{cubuk_randaugment_2020,
	title = {{RandAugment}: {Practical} automated data augmentation with a reduced search space},
	shorttitle = {Randaugment},
	url = {https://ieeexplore.ieee.org/document/9150790},
	doi = {10.1109/CVPRW50498.2020.00359},
	urldate = {2024-05-11},
	booktitle = {2020 {IEEE}/{CVF} {Conference} on {Computer} {Vision} and {Pattern} {Recognition} {Workshops} ({CVPRW})},
	author = {Cubuk, Ekin D. and Zoph, Barret and Shlens, Jonathon and Le, Quoc V.},
	month = jun,
	year = {2020},
	note = {ISSN: 2160-7516},
	pages = {3008--3017},
}

@inproceedings{parhar_hyperionsolarnet_2021,
	title = {{HyperionSolarNet}: {Solar} {Panel} {Detection} from {Aerial} {Images}},
	url = {https://www.climatechange.ai/papers/neurips2021/41},
	booktitle = {{NeurIPS} 2021 {Workshop} on {Tackling} {Climate} {Change} with {Machine} {Learning}},
	author = {Parhar, Poonam and Sawasaki, Ryan and Todeschini, Alberto and Reed, Colorado and Vahabi, Hossein and Nusaputra, Nathan and Vergara, Felipe},
	year = {2021},
}

@inproceedings{rausch_enriched_2020,
	title = {An {Enriched} {Automated} {PV} {Registry}: {Combining} {Image} {Recognition} and {3D} {Building} {Data}},
	url = {https://www.climatechange.ai/papers/neurips2020/46},
	booktitle = {{NeurIPS} 2020 {Workshop} on {Tackling} {Climate} {Change} with {Machine} {Learning}},
	author = {Rausch, Benjamin and Mayer, Kevin and Arlt, Marie-Louise and Gust, Gunther and Staudt, Philipp and Weinhardt, Christof and Neumann, Dirk and Rajagopal, Ram},
	year = {2020},
}

@inproceedings{hendrycks_augmix_2020,
	title = {{AugMix}: {A} {Simple} {Data} {Processing} {Method} to {Improve} {Robustness} and {Uncertainty}},
	url = {https://openreview.net/forum?id=S1gmrxHFvB},
	booktitle = {8th {International} {Conference} on {Learning} {Representations}, {ICLR} 2020, {Addis} {Ababa}, {Ethiopia}, {April} 26-30, 2020},
	publisher = {OpenReview.net},
	author = {Hendrycks, Dan and Mu, Norman and Cubuk, Ekin Dogus and Zoph, Barret and Gilmer, Justin and Lakshminarayanan, Balaji},
	year = {2020},
}

@inproceedings{kasmi_assessment_2023,
	title = {Assessment of the {Reliablity} of a {Model}'s {Decision} by {Generalizing} {Attribution} to the {Wavelet} {Domain}},
	copyright = {All rights reserved},
	url = {http://arxiv.org/abs/2305.14979},
	doi = {10.48550/arXiv.2305.14979},
	urldate = {2023-09-25},
	booktitle = {{XAI} in {Action}: {Past}, {Present}, and {Future} {Applications} workshop at {NeurIPS} 2023},
	publisher = {arXiv},
	author = {Kasmi, Gabriel and Dubus, Laurent and Saint-Drenan, Yves-Marie and Blanc, Philippe},
	month = sep,
	year = {2023},
	note = {arXiv:2305.14979 [cs, stat]},
}

@article{sun_shift_2022,
	title = {{SHIFT}: {A} {Synthetic} {Driving} {Dataset} for {Continuous} {Multi}-{Task} {Domain} {Adaptation}},
	shorttitle = {{SHIFT}},
	url = {https://openaccess.thecvf.com/content/CVPR2022/html/Sun_SHIFT_A_Synthetic_Driving_Dataset_for_Continuous_Multi-Task_Domain_Adaptation_CVPR_2022_paper.html},
	language = {en},
	urldate = {2023-11-06},
	journal = {Proceedings of the IEEE/CVF Conference on Computer Vision and Pattern Recognition (CVPR)},
	author = {Sun, Tao and Segu, Mattia and Postels, Janis and Wang, Yuxuan and Van Gool, Luc and Schiele, Bernt and Tombari, Federico and Yu, Fisher},
	year = {2022},
	pages = {21371--21382},
}

@article{hendrycks_pixmix_2022,
	title = {{PixMix}: {Dreamlike} {Pictures} {Comprehensively} {Improve} {Safety} {Measures}},
	shorttitle = {{PixMix}},
	url = {http://arxiv.org/abs/2112.05135},
	doi = {10.48550/arXiv.2112.05135},
	urldate = {2023-04-18},
	journal = {Proceedings of the IEEE/CVF Conference on Computer Vision and Pattern Recognition (CVPR)},
	author = {Hendrycks, Dan and Zou, Andy and Mazeika, Mantas and Tang, Leonard and Li, Bo and Song, Dawn and Steinhardt, Jacob},
	month = mar,
	year = {2022},
	note = {arXiv:2112.05135 [cs]},
}

@article{pierro_impact_2022,
	title = {Impact of {PV}/{Wind} {Forecast} {Accuracy} and {National} {Transmission} {Grid} {Reinforcement} on the {Italian} {Electric} {System}},
	volume = {15},
	issn = {1996-1073},
	url = {https://www.mdpi.com/1996-1073/15/23/9086},
	doi = {10.3390/en15239086},
	language = {en},
	number = {23},
	urldate = {2022-12-08},
	journal = {Energies},
	author = {Pierro, Marco and Liolli, Fabio Romano and Gentili, Damiano and Petitta, Marcello and Perez, Richard and Moser, David and Cornaro, Cristina},
	month = nov,
	year = {2022},
	pages = {9086},
}

@misc{flora_comparing_2022,
	title = {Comparing {Explanation} {Methods} for {Traditional} {Machine} {Learning} {Models} {Part} 1: {An} {Overview} of {Current} {Methods} and {Quantifying} {Their} {Disagreement}},
	shorttitle = {Comparing {Explanation} {Methods} for {Traditional} {Machine} {Learning} {Models} {Part} 1},
	url = {http://arxiv.org/abs/2211.08943},
	language = {en},
	urldate = {2024-03-16},
	publisher = {arXiv},
	author = {Flora, Montgomery and Potvin, Corey and McGovern, Amy and Handler, Shawn},
	month = nov,
	year = {2022},
	note = {arXiv:2211.08943 [physics, stat]},
}

@article{yang_weakly-semi_2024,
	title = {Weakly-semi supervised extraction of rooftop photovoltaics from high-resolution images based on segment anything model and class activation map},
	volume = {361},
	issn = {03062619},
	url = {https://linkinghub.elsevier.com/retrieve/pii/S0306261924003477},
	doi = {10.1016/j.apenergy.2024.122964},
	language = {en},
	urldate = {2024-03-12},
	journal = {Applied Energy},
	author = {Yang, Ruiqing and He, Guojin and Yin, Ranyu and Wang, Guizhou and Zhang, Zhaoming and Long, Tengfei and Peng, Yan},
	month = may,
	year = {2024},
	pages = {122964},
}

@article{zhang_grad-cam_2021,
	title = {Grad-{CAM} helps interpret the deep learning models trained to classify multiple sclerosis types using clinical brain magnetic resonance imaging},
	volume = {353},
	issn = {01650270},
	url = {https://linkinghub.elsevier.com/retrieve/pii/S0165027021000339},
	doi = {10.1016/j.jneumeth.2021.109098},
	language = {en},
	urldate = {2024-01-18},
	journal = {Journal of Neuroscience Methods},
	author = {Zhang, Yunyan and Hong, Daphne and McClement, Daniel and Oladosu, Olayinka and Pridham, Glen and Slaney, Garth},
	month = apr,
	year = {2021},
	pages = {109098},
}

@misc{ign_bd_2024,
	title = {{BD} {ORTHO}® {\textbar} {Géoservices}},
	url = {https://geoservices.ign.fr/bdortho},
	urldate = {2024-01-18},
	author = {{IGN}},
	year = {2024},
}

@techreport{rte_france_energy_2022,
	title = {Energy {Pathways} to 2050},
	url = {https://rte-futursenergetiques2050.com/},
	urldate = {2023-09-07},
	institution = {RTE France},
	author = {{RTE France}},
	year = {2022},
}

@misc{bommasani_opportunities_2022,
	title = {On the {Opportunities} and {Risks} of {Foundation} {Models}},
	url = {http://arxiv.org/abs/2108.07258},
	language = {en},
	urldate = {2023-12-16},
	publisher = {arXiv},
	author = {Bommasani, Rishi and Hudson, Drew A. and Adeli, Ehsan and Altman, Russ and Arora, Simran and von Arx, Sydney and Bernstein, Michael S. and Bohg, Jeannette and Bosselut, Antoine and Brunskill, Emma and Brynjolfsson, Erik and Buch, Shyamal and Card, Dallas and Castellon, Rodrigo and Chatterji, Niladri and Chen, Annie and Creel, Kathleen and Davis, Jared Quincy and Demszky, Dora and Donahue, Chris and Doumbouya, Moussa and Durmus, Esin and Ermon, Stefano and Etchemendy, John and Ethayarajh, Kawin and Fei-Fei, Li and Finn, Chelsea and Gale, Trevor and Gillespie, Lauren and Goel, Karan and Goodman, Noah and Grossman, Shelby and Guha, Neel and Hashimoto, Tatsunori and Henderson, Peter and Hewitt, John and Ho, Daniel E. and Hong, Jenny and Hsu, Kyle and Huang, Jing and Icard, Thomas and Jain, Saahil and Jurafsky, Dan and Kalluri, Pratyusha and Karamcheti, Siddharth and Keeling, Geoff and Khani, Fereshte and Khattab, Omar and Koh, Pang Wei and Krass, Mark and Krishna, Ranjay and Kuditipudi, Rohith and Kumar, Ananya and Ladhak, Faisal and Lee, Mina and Lee, Tony and Leskovec, Jure and Levent, Isabelle and Li, Xiang Lisa and Li, Xuechen and Ma, Tengyu and Malik, Ali and Manning, Christopher D. and Mirchandani, Suvir and Mitchell, Eric and Munyikwa, Zanele and Nair, Suraj and Narayan, Avanika and Narayanan, Deepak and Newman, Ben and Nie, Allen and Niebles, Juan Carlos and Nilforoshan, Hamed and Nyarko, Julian and Ogut, Giray and Orr, Laurel and Papadimitriou, Isabel and Park, Joon Sung and Piech, Chris and Portelance, Eva and Potts, Christopher and Raghunathan, Aditi and Reich, Rob and Ren, Hongyu and Rong, Frieda and Roohani, Yusuf and Ruiz, Camilo and Ryan, Jack and Ré, Christopher and Sadigh, Dorsa and Sagawa, Shiori and Santhanam, Keshav and Shih, Andy and Srinivasan, Krishnan and Tamkin, Alex and Taori, Rohan and Thomas, Armin W. and Tramèr, Florian and Wang, Rose E. and Wang, William and Wu, Bohan and Wu, Jiajun and Wu, Yuhuai and Xie, Sang Michael and Yasunaga, Michihiro and You, Jiaxuan and Zaharia, Matei and Zhang, Michael and Zhang, Tianyi and Zhang, Xikun and Zhang, Yuhui and Zheng, Lucia and Zhou, Kaitlyn and Liang, Percy},
	month = jul,
	year = {2022},
	note = {arXiv:2108.07258 [cs]},
}

@article{freitas_artificial_2023,
	title = {An {Artificial} {Intelligence}-{Based} {Framework} to {Accelerate} {Data}-{Driven} {Policies} to {Promote} {Solar} {Photovoltaics} in {Lisbon}},
	volume = {n/a},
	copyright = {© 2023 The Authors. Solar RRL published by Wiley-VCH GmbH},
	issn = {2367-198X},
	url = {https://onlinelibrary.wiley.com/doi/abs/10.1002/solr.202300597},
	doi = {10.1002/solr.202300597},
	language = {en},
	number = {n/a},
	urldate = {2023-11-06},
	journal = {Solar RRL},
	author = {Freitas, Sara and Silva, Miguel and Silva, Eduardo and Marceddu, Alessandro and Miccoli, Massimo and Gnatyuk, Petro and Marangoni, Luca and Amicone, Alessandro},
	year = {2023},
	note = {\_eprint: https://onlinelibrary.wiley.com/doi/pdf/10.1002/solr.202300597},
	pages = {2300597},
}

@incollection{avidan_spectral_2022,
	address = {Cham},
	title = {A {Spectral} {View} of {Randomized} {Smoothing} {Under} {Common} {Corruptions}: {Benchmarking} and {Improving} {Certified} {Robustness}},
	volume = {13664},
	isbn = {978-3-031-19771-0 978-3-031-19772-7},
	shorttitle = {A {Spectral} {View} of {Randomized} {Smoothing} {Under} {Common} {Corruptions}},
	url = {https://link.springer.com/10.1007/978-3-031-19772-7_38},
	language = {en},
	urldate = {2023-11-17},
	booktitle = {Computer {Vision} – {ECCV} 2022},
	publisher = {Springer Nature Switzerland},
	author = {Sun, Jiachen and Mehra, Akshay and Kailkhura, Bhavya and Chen, Pin-Yu and Hendrycks, Dan and Hamm, Jihun and Mao, Z. Morley},
	editor = {Avidan, Shai and Brostow, Gabriel and Cissé, Moustapha and Farinella, Giovanni Maria and Hassner, Tal},
	year = {2022},
	doi = {10.1007/978-3-031-19772-7_38},
	note = {Series Title: Lecture Notes in Computer Science},
	pages = {654--671},
}

@inproceedings{pooch_can_2020,
	address = {Cham},
	series = {Lecture {Notes} in {Computer} {Science}},
	title = {Can {We} {Trust} {Deep} {Learning} {Based} {Diagnosis}? {The} {Impact} of {Domain} {Shift} in {Chest} {Radiograph} {Classification}},
	isbn = {978-3-030-62469-9},
	shorttitle = {Can {We} {Trust} {Deep} {Learning} {Based} {Diagnosis}?},
	doi = {10.1007/978-3-030-62469-9_7},
	language = {en},
	booktitle = {Thoracic {Image} {Analysis}},
	publisher = {Springer International Publishing},
	author = {Pooch, Eduardo H. P. and Ballester, Pedro and Barros, Rodrigo C.},
	editor = {Petersen, Jens and San José Estépar, Raúl and Schmidt-Richberg, Alexander and Gerard, Sarah and Lassen-Schmidt, Bianca and Jacobs, Colin and Beichel, Reinhard and Mori, Kensaku},
	year = {2020},
	pages = {74--83},
}

@inproceedings{kasmi_towards_2022,
	series = {{CEUR} {Workshop} {Proceedings}},
	title = {Towards {Unsupervised} {Assessment} with {Open}-{Source} {Data} of the {Accuracy} of {Deep} {Learning}-{Based} {Distributed} {PV} {Mapping}},
	volume = {3343},
	copyright = {All rights reserved},
	url = {https://ceur-ws.org/Vol-3343/paper6.pdf},
	booktitle = {Proceedings of {MACLEAN}: {MAChine} {Learning} for {EArth} {ObservatioN} {Workshop} co-located with the {European} {Conference} on {Machine} {Learning} and {Principles} and {Practice} of {Knowledge} {Discovery} in {Databases} ({ECML}/{PKDD} 2022), {Grenoble}, {France}, {September} 18-22, 2022},
	publisher = {CEUR-WS.org},
	author = {Kasmi, Gabriel and Dubus, Laurent and Saint-Drenan, Yves-Marie and Blanc, Philippe},
	editor = {Corpetti, Thomas and Ienco, Dino and Interdonato, Roberto and Pham, Minh-Tan and Lefèvre, Sébastien},
	year = {2022},
}

@phdthesis{kausika_gis4pv_2022,
	title = {{GIS4PV}: {A} technological impact assessment of the application of {GIS} for {Photovoltaic} {Solar} {Energy}.},
	shorttitle = {{GIS4PV}},
	url = {https://dspace.library.uu.nl/handle/1874/420472},
	language = {en},
	urldate = {2023-09-14},
	school = {Utrecht University},
	author = {Kausika, Bala Bhavya},
	month = may,
	year = {2022},
	doi = {10.33540/1371},
}

@article{lindahl_mapping_2023,
	title = {Mapping of decentralised photovoltaic and solar thermal systems by remote sensing aerial imagery and deep machine learning for statistic generation},
	issn = {2666-5468},
	url = {https://www.sciencedirect.com/science/article/pii/S2666546823000721},
	doi = {10.1016/j.egyai.2023.100300},
	urldate = {2023-09-14},
	journal = {Energy and AI},
	author = {Lindahl, Johan and Johansson, Robert and Lingfors, David},
	month = sep,
	year = {2023},
	pages = {100300},
}

@misc{achtibat_where_2022,
	title = {From "{Where}" to "{What}": {Towards} {Human}-{Understandable} {Explanations} through {Concept} {Relevance} {Propagation}},
	shorttitle = {From "{Where}" to "{What}"},
	url = {http://arxiv.org/abs/2206.03208},
	doi = {10.48550/arXiv.2206.03208},
	urldate = {2023-09-12},
	publisher = {arXiv},
	author = {Achtibat, Reduan and Dreyer, Maximilian and Eisenbraun, Ilona and Bosse, Sebastian and Wiegand, Thomas and Samek, Wojciech and Lapuschkin, Sebastian},
	month = jun,
	year = {2022},
	note = {arXiv:2206.03208 [cs]},
}

@article{frimane_identifying_2023,
	title = {Identifying small decentralized solar systems in aerial images using deep learning},
	volume = {262},
	issn = {0038-092X},
	url = {https://www.sciencedirect.com/science/article/pii/S0038092X23004474},
	doi = {10.1016/j.solener.2023.111822},
	language = {en},
	urldate = {2023-07-18},
	journal = {Solar Energy},
	author = {Frimane, Âzeddine and Johansson, Robert and Munkhammar, Joakim and Lingfors, David and Lindahl, Johan},
	month = sep,
	year = {2023},
	pages = {111822},
}

@article{gorelick_google_2017,
	title = {Google {Earth} {Engine}: {Planetary}-scale geospatial analysis for everyone},
	volume = {202},
	journal = {Remote sensing of Environment},
	author = {Gorelick, Noel and Hancher, Matt and Dixon, Mike and Ilyushchenko, Simon and Thau, David and Moore, Rebecca},
	year = {2017},
	note = {Publisher: Elsevier},
	pages = {18--27},
}

@article{arnaudo_comparative_2023,
	title = {A {Comparative} {Evaluation} of {Deep} {Learning} {Techniques} for {Photovoltaic} {Panel} {Detection} from {Aerial} {Images}},
	issn = {2169-3536},
	url = {https://ieeexplore.ieee.org/document/10122915/},
	doi = {10.1109/ACCESS.2023.3275435},
	language = {en},
	urldate = {2023-05-17},
	journal = {IEEE Access},
	author = {Arnaudo, Edoardo and Blanco, Giacomo and Monti, Antonino and Bianco, Gabriele and Monaco, Cristina and Pasquali, Paolo and Dominici, Fabrizio},
	year = {2023},
	pages = {1--1},
}

@article{jansen_analysis_1999,
	title = {Analysis of variance designs for model output},
	volume = {117},
	issn = {0010-4655},
	url = {https://www.sciencedirect.com/science/article/pii/S0010465598001544},
	doi = {10.1016/S0010-4655(98)00154-4},
	language = {en},
	number = {1},
	urldate = {2023-05-06},
	journal = {Computer Physics Communications},
	author = {Jansen, Michiel J. W.},
	month = mar,
	year = {1999},
	pages = {35--43},
}

@article{mallat_theory_1989,
	title = {A theory for multiresolution signal decomposition: the wavelet representation},
	volume = {11},
	issn = {1939-3539},
	shorttitle = {A theory for multiresolution signal decomposition},
	doi = {10.1109/34.192463},
	number = {7},
	journal = {IEEE Transactions on Pattern Analysis and Machine Intelligence},
	author = {Mallat, S.G.},
	month = jul,
	year = {1989},
	note = {Conference Name: IEEE Transactions on Pattern Analysis and Machine Intelligence},
	pages = {674--693},
}

@article{kasmi_crowdsourced_2023,
	title = {A crowdsourced dataset of aerial images with annotated solar photovoltaic arrays and installation metadata},
	volume = {10},
	copyright = {All rights reserved},
	issn = {2052-4463},
	url = {https://doi.org/10.1038/s41597-023-01951-4},
	doi = {10.1038/s41597-023-01951-4},
	number = {1},
	journal = {Scientific Data},
	author = {Kasmi, Gabriel and Saint-Drenan, Yves-Marie and Trebosc, David and Jolivet, Raphaël and Leloux, Jonathan and Sarr, Babacar and Dubus, Laurent},
	month = jan,
	year = {2023},
	pages = {59},
}

@inproceedings{malof_automatic_2015,
	address = {Palermo, Italy},
	title = {Automatic solar photovoltaic panel detection in satellite imagery},
	isbn = {978-1-4799-9982-8},
	url = {http://ieeexplore.ieee.org/document/7418643/},
	doi = {10.1109/ICRERA.2015.7418643},
	language = {en},
	urldate = {2022-12-08},
	booktitle = {2015 {International} {Conference} on {Renewable} {Energy} {Research} and {Applications} ({ICRERA})},
	publisher = {IEEE},
	author = {Malof, Jordan M. and {Rui Hou} and Collins, Leslie M. and Bradbury, Kyle and Newell, Richard},
	month = nov,
	year = {2015},
	pages = {1428--1431},
}

@article{yu_deepsolar_2018,
	title = {{DeepSolar}: {A} {Machine} {Learning} {Framework} to {Efficiently} {Construct} a {Solar} {Deployment} {Database} in the {United} {States}},
	volume = {2},
	issn = {2542-4351},
	shorttitle = {{DeepSolar}},
	url = {https://www.sciencedirect.com/science/article/pii/S2542435118305701},
	doi = {10.1016/j.joule.2018.11.021},
	language = {en},
	number = {12},
	urldate = {2022-12-08},
	journal = {Joule},
	author = {Yu, Jiafan and Wang, Zhecheng and Majumdar, Arun and Rajagopal, Ram},
	month = dec,
	year = {2018},
	pages = {2605--2617},
}

@inproceedings{yuan_large-scale_2016,
	address = {Washington DC,USA},
	title = {Large-scale solar panel mapping from aerial images using deep convolutional networks},
	isbn = {978-1-4673-9005-7},
	url = {http://ieeexplore.ieee.org/document/7840915/},
	doi = {10.1109/BigData.2016.7840915},
	language = {en},
	urldate = {2022-12-08},
	booktitle = {2016 {IEEE} {International} {Conference} on {Big} {Data} ({Big} {Data})},
	publisher = {IEEE},
	author = {Yuan, Jiangye and Yang, Hsiu-Han Lexie and Omitaomu, Olufemi A. and Bhaduri, Budhendra L.},
	month = dec,
	year = {2016},
	pages = {2703--2708},
}

@article{malof_automatic_2016,
	title = {Automatic detection of solar photovoltaic arrays in high resolution aerial imagery},
	volume = {183},
	issn = {03062619},
	url = {https://linkinghub.elsevier.com/retrieve/pii/S0306261916313009},
	doi = {10.1016/j.apenergy.2016.08.191},
	language = {en},
	urldate = {2022-12-08},
	journal = {Applied Energy},
	author = {Malof, Jordan M. and Bradbury, Kyle and Collins, Leslie M. and Newell, Richard G.},
	month = dec,
	year = {2016},
	pages = {229--240},
}

@article{lapuschkin_unmasking_2019,
	title = {Unmasking clever hans predictors and assessing what machines really learn},
	volume = {10},
	number = {1},
	journal = {Nature communications},
	author = {Lapuschkin, Sebastian and Wäldchen, Stephan and Binder, Alexander and Montavon, Grégoire and Samek, Wojciech and Müller, Klaus-Robert},
	year = {2019},
	note = {Publisher: Nature Publishing Group},
	pages = {1--8},
}

@inproceedings{zech_predicting_2020,
	address = {Calgary, AB, Canada},
	title = {Predicting {PV} {Areas} in {Aerial} {Images} with {Deep} {Learning}},
	isbn = {978-1-72816-115-0},
	url = {https://ieeexplore.ieee.org/document/9300636/},
	doi = {10.1109/PVSC45281.2020.9300636},
	language = {en},
	urldate = {2022-12-08},
	booktitle = {2020 47th {IEEE} {Photovoltaic} {Specialists} {Conference} ({PVSC})},
	publisher = {IEEE},
	author = {Zech, Matthias and Ranalli, Joseph},
	month = jun,
	year = {2020},
	pages = {0767--0774},
}

@inproceedings{wang_poor_2017,
	address = {Washington, DC},
	title = {The poor generalization of deep convolutional networks to aerial imagery from new geographic locations: an empirical study with solar array detection},
	isbn = {978-1-5386-1235-4},
	shorttitle = {The poor generalization of deep convolutional networks to aerial imagery from new geographic locations},
	url = {https://ieeexplore.ieee.org/document/8457965/},
	doi = {10.1109/AIPR.2017.8457965},
	language = {en},
	urldate = {2022-12-08},
	booktitle = {2017 {IEEE} {Applied} {Imagery} {Pattern} {Recognition} {Workshop} ({AIPR})},
	publisher = {IEEE},
	author = {Wang, Rui and Camilo, Joseph and Collins, Leslie M. and Bradbury, Kyle and Malof, Jordan M.},
	month = oct,
	year = {2017},
	pages = {1--8},
}

@inproceedings{mayer_deepsolar_2020,
	address = {Istanbul, Turkey},
	title = {{DeepSolar} for {Germany}: {A} deep learning framework for {PV} system mapping from aerial imagery},
	isbn = {978-1-72814-701-7},
	shorttitle = {{DeepSolar} for {Germany}},
	url = {https://ieeexplore.ieee.org/document/9203258/},
	doi = {10.1109/SEST48500.2020.9203258},
	language = {en},
	urldate = {2022-12-08},
	booktitle = {2020 {International} {Conference} on {Smart} {Energy} {Systems} and {Technologies} ({SEST})},
	publisher = {IEEE},
	author = {Mayer, Kevin and Wang, Zhecheng and Arlt, Marie-Louise and Neumann, Dirk and Rajagopal, Ram},
	month = sep,
	year = {2020},
	pages = {1--6},
}

@article{li_understanding_2021,
	title = {Understanding rooftop {PV} panel semantic segmentation of satellite and aerial images for better using machine learning},
	volume = {4},
	issn = {26667924},
	url = {https://linkinghub.elsevier.com/retrieve/pii/S2666792421000494},
	doi = {10.1016/j.adapen.2021.100057},
	language = {en},
	urldate = {2022-12-08},
	journal = {Advances in Applied Energy},
	author = {Li, Peiran and Zhang, Haoran and Guo, Zhiling and Lyu, Suxing and Chen, Jinyu and Li, Wenjing and Song, Xuan and Shibasaki, Ryosuke and Yan, Jinyue},
	month = nov,
	year = {2021},
	pages = {100057},
}

@inproceedings{golovko_development_2018,
	address = {Kharkiv, Ukraine},
	title = {Development of {Solar} {Panels} {Detector}},
	isbn = {978-1-5386-6609-8 978-1-5386-6611-1},
	url = {https://ieeexplore.ieee.org/document/8632132/},
	doi = {10.1109/INFOCOMMST.2018.8632132},
	language = {en},
	urldate = {2022-12-08},
	booktitle = {2018 {International} {Scientific}-{Practical} {Conference} {Problems} of {Infocommunications}. {Science} and {Technology} ({PIC} {S}\&{T})},
	publisher = {IEEE},
	author = {Golovko, Vladimir and Kroshchanka, Alexander and Bezobrazov, Sergei and Sachenko, Anatoliy and Komar, Myroslav and Novosad, Oleksandr},
	month = oct,
	year = {2018},
	pages = {761--764},
}

@book{mallat_wavelet_1999,
	title = {A wavelet tour of signal processing},
	publisher = {Elsevier},
	author = {Mallat, Stéphane},
	year = {1999},
}

@book{vapnik_nature_1999,
	title = {The nature of statistical learning theory},
	publisher = {Springer science \& business media},
	author = {Vapnik, Vladimir},
	year = {1999},
}

@inproceedings{murray_zoom_2019,
	title = {Zoom {In}, {Zoom} {Out}: {Injecting} {Scale} {Invariance} into {Landuse} {Classification} {CNNs}},
	booktitle = {{IGARSS} 2019-2019 {IEEE} {International} {Geoscience} and {Remote} {Sensing} {Symposium}},
	publisher = {IEEE},
	author = {Murray, Jesse and Marcos, Diego and Tuia, Devis},
	year = {2019},
	pages = {5240--5243},
}

@article{kausika_geoai_2021,
	title = {{GeoAI} for detection of solar photovoltaic installations in the {Netherlands}},
	volume = {6},
	issn = {26665468},
	url = {https://linkinghub.elsevier.com/retrieve/pii/S2666546821000604},
	doi = {10.1016/j.egyai.2021.100111},
	language = {en},
	urldate = {2022-12-08},
	journal = {Energy and AI},
	author = {Kausika, Bala Bhavya and Nijmeijer, Diede and Reimerink, Iris and Brouwer, Peter and Liem, Vera},
	month = dec,
	year = {2021},
	pages = {100111},
}

@article{mayer_3d-pv-locator_2022,
	title = {{3D}-{PV}-{Locator}: {Large}-scale detection of rooftop-mounted photovoltaic systems in {3D}},
	volume = {310},
	issn = {03062619},
	shorttitle = {{3D}-{PV}-{Locator}},
	url = {https://linkinghub.elsevier.com/retrieve/pii/S0306261921016937},
	doi = {10.1016/j.apenergy.2021.118469},
	language = {en},
	urldate = {2022-12-08},
	journal = {Applied Energy},
	author = {Mayer, Kevin and Rausch, Benjamin and Arlt, Marie-Louise and Gust, Gunther and Wang, Zhecheng and Neumann, Dirk and Rajagopal, Ram},
	month = mar,
	year = {2022},
	pages = {118469},
}

@article{haegel_terawatt-scale_2017,
	title = {Terawatt-scale photovoltaics: {Trajectories} and challenges},
	volume = {356},
	number = {6334},
	journal = {Science},
	author = {Haegel, Nancy M and Margolis, Robert and Buonassisi, Tonio and Feldman, David and Froitzheim, Armin and Garabedian, Raffi and Green, Martin and Glunz, Stefan and Henning, Hans-Martin and Holder, Burkhard and {others}},
	year = {2017},
	note = {Publisher: American Association for the Advancement of Science},
	pages = {141--143},
}

@inproceedings{he_deep_2016,
	title = {Deep residual learning for image recognition},
	booktitle = {Proceedings of the {IEEE} conference on computer vision and pattern recognition},
	author = {He, Kaiming and Zhang, Xiangyu and Ren, Shaoqing and Sun, Jian},
	year = {2016},
	pages = {770--778},
}
